\pdfoutput=1

\documentclass[11pt]{article}

\usepackage[final]{acl}

\usepackage{times}
\usepackage{latexsym}

\usepackage[T1]{fontenc}

\usepackage[utf8]{inputenc}

\usepackage{microtype}

\usepackage{inconsolata}

\usepackage{graphicx}
\usepackage{pifont}
\usepackage{graphicx}
\usepackage{amsmath}
\usepackage{amssymb}
\usepackage{hyperref}[colorlinks,linkcolor=blue]
\usepackage{float} 
\usepackage{subfigure} 
\usepackage{booktabs,makecell, multirow, tabularx}
\usepackage{algorithm}
\usepackage{algorithmic}  
\usepackage[edges]{forest}
\usetikzlibrary{shadows}
\usepackage{bbding}
\usepackage[bottom]{footmisc}
\usepackage{tablefootnote}

\usepackage[normalem]{ulem}
\usepackage{color}
\usepackage{array} 
\useunder{\uline}{\ul}{}
\usepackage{arydshln}
\newcommand{\Rmnum}[1]{\uppercase\expandafter{\romannumeral #1}} 
 
\title{Towards Concise and Adaptive Thinking in Large Reasoning Models: A Survey}

\author{Jason Zhu$^{*}$, Hongyu Li\thanks{Equal contribution} \\
\texttt{\{jszjg1991,hongyuli102799\}@gmail.com}
}

\begin{document}

\maketitle

\begin{abstract}
Large reasoning models (LRMs) like OpenAI o1 and DeepSeek R1 have demonstrated impressive performance on complex reasoning tasks like mathematics and programming with long Chain-of-Thought (CoT) reasoning sequences (slow-thinking), compared with traditional large language models (fast-thinking). However, these reasoning models also face a huge challenge that generating unnecessarily lengthy and redundant reasoning chains even for trivial questions. This phenomenon leads to a significant waste of inference resources, increases the response time for simple queries, and hinders the practical application of LRMs in real-world products. To this end, it is crucial to shorten lengthy reasoning chains and learn adaptive reasoning between fast and slow thinking based on input difficulty. In this survey, we provide a comprehensive overview of recent progress in concise and adaptive thinking for efficient reasoning of LRMs, including methodologies, benchmarks, and challenges for future exploration. We hope this survey can help researchers quickly understand the landscape of this field and inspire novel adaptive thinking ideas to facilitate better usage of LRMs.
\end{abstract}

\section{Introduction}
Recent advances~\cite{openai2024openaio1card,deepseekai2025deepseekr1incentivizingreasoningcapability,seed2025seed15thinkingadvancingsuperbreasoning,ji2025amthinkingv1advancingfrontierreasoning,bercovich2025llamanemotronefficientreasoningmodels,coreteam2025mimounlockingreasoningpotential,he2025skyworkopenreasoner1,zhang2025100daysdeepseekr1survey,minimax2025minimaxm1scalingtesttimecompute} of large reasoning models (LRMs) have demonstrated amazing capabilities in complex reasoning domains, especially in Olympiad-level mathematics and programming tasks. Compared with traditional large language models~(LLMs, e.g., LLaMA~\citep{grattafiori2024llama3herdmodels}, Qwen2.5~\citep{qwen2025qwen25technicalreport}), these LRMs gain such significant improvements by explicitly generating long Chain-of-Thoughts (CoT). Such long CoT capability with step-by-step analysis and self-reflection characteristic is recognized as the core technique of achieving human-level intelligence by shifting the model's thinking mode from intuitive and fast decisions (System 1) to slower and deliberate thinking (System 2)~\cite{li202512surveyreasoning,pan2025surveyslowthinkingbasedreasoning}.

Meanwhile, a potential issue behind LRMs is the tendency to generate lengthy and redundant content that dramatically increases the inference cost and hurts the user experience, which is also widely known as the \textbf{overthinking} phenomenon~\cite{chen2025think23overthinkingo1like,fan2025missingpremiseexacerbatesoverthinking}. Many recent studies~\cite{zeng2025revisitingtesttimescalingo1like,su2025underthinkingoverthinkingempiricalstudy,zhao2025tradeoffslargereasoningmodels,yang2025thinkingoptimalscalingtesttimecompute,jin2024impactreasoningsteplength,wu2025lessunderstandingchainofthoughtlength,lee2025llmscompresschainofthoughttoken,ghosal2025doesthinkinghelpunderstanding,sun2025empiricalstudyllmreasoning,hassid2025dontoverthinkitpreferring} try to understand how generation length affects the accuracy of LRMs and reveal that deliberative reasoning capability by extremely long reasoning chains does not consistently improve model performance across diverse tasks. There exist optimal reasoning lengths for different tasks that allow the model to think adaptively based on input difficulty. To this end, either shortening the verbose and lengthy CoT to more concise (\textbf{Concise Thinking}) or adaptively triggering slow thinking based on input difficulty (\textbf{Adaptive Thinking}) is critical to reduce the inference cost of LRMs, thus making these models more accessible in real-world applications. Table~\ref{table:release_model} provides a summary of selected released models with adaptive thinking capability.

In this survey, we aim to provide a comprehensive overview of recent cutting-edge research in concise and adaptive thinking of LRMs, including the shortening and refinement of long CoTs, as well as the adaptive toggle of thinking modes based on input difficulty. As illustrated in Figure~\ref{fig:taxonomy}, we categorize existing works into two directions: (1) \textbf{Training-free methods}, which aim to achieve concise and adaptive thinking without any training effort, including prompt guided, pipeline based, decoding manipulation, and model merging. (2) \textbf{Training-based methods}, which focus on shortening the reasoning length and teaching the LRMs to think adaptively by fine-tuning (SFT/DPO) or reinforcement learning~(RL).

\noindent\textbf{What's not covered?} Besides adaptive thinking, there are also many other approaches for efficient reasoning of LRMs, including: (1) small language model distillation~\cite{magister2023teachingsmalllanguagemodels,paliotta2025thinkingslowfastscaling,jiang2025drpdistilledreasoningpruning}, quantization~\cite{liu2025quantizationhurtsreasoningempirical}, and speculative decoding~\cite{fu2025r2refficientlynavigatingdivergent}, which designs to build a much more light-weighted model that maintains similar performance to the original one while achieving faster inference speed. (2) model architecture optimization~\cite{wang2025m1scalabletesttimecompute,minimax2025minimaxm1scalingtesttimecompute}, which is dedicated to designing more inference-friendly blocks to reduce computational cost, including linear attention, sparse attention, etc. (3) efficient reasoning in multimodal models~\cite{xiao2025fastslowthinkinglargevisionlanguage}, which aims to balance fast-slow thinking in multimodal models. This survey will not cover these topics.

There have been several surveys~\cite{qu2025surveyefficientreasoninglarge,sui2025stopoverthinkingsurveyefficient,liu2025efficientinferencelargereasoning,feng2025efficientreasoningmodelssurvey,alomrani2025reasoningbudgetsurveyadaptive} about efficient reasoning of LRMs. Compared with these works, we focus on concise and adaptive thinking, which is essential to build a unified model that possesses both strong chat and reasoning capabilities. We conduct a comprehensive review of the latest progress in this field, including technical approaches, data and evaluation methodologies. From a practical perspective, we systematically categorize and analyze existing solutions. Finally, we identify key limitations in current field and promising future directions for exploration.

\begin{table*}[!ht]
\centering
\resizebox{0.7\textwidth}{!}{
\begin{tabular}{lc}
\hline
\multicolumn{1}{c}{\textbf{Name}}                       & \multicolumn{1}{c}{\textbf{Organization}} \\ \hline
Claude 3.7 Sonnet~\cite{claude37sonnet}                                         & Anthropic                    \\ \hline
Qwen3~\cite{yang2025qwen3technicalreport}                                         & Alibaba                    \\ \hline
Llama-Nemotron\cite{bercovich2025llamanemotronefficientreasoningmodels}                                         & Nvidia               \\ \hline
iFLYTEK Spark~\cite{xinghuo2025x1model}   & iFLYTEK   \\ \hline
KwaiCoder-AutoThink-preview~\cite{KwaiPilot2025AutoThink}   & Kuaishou \\ \hline
doubao-seed-1.6~\cite{seed2025adacotmodel} & Bytedance \\ \hline
Hunyuan-TurboS~\cite{tencenthunyuanteam2025hunyuanturbosadvancinglargelanguage} & Tencent \\ \hline
Hunyuan-A13B~\cite{hunyuana13b} & Tencent \\ \hline
\end{tabular}}
\caption{Summary of Selected Adaptive/Hybrid Reasoning Models.}
\label{table:release_model}
\end{table*}

\definecolor{trainfreeColor}{HTML}{f95738}  
\definecolor{trainColor}{HTML}{bdd5ea}      
\definecolor{hidden-red}{RGB}{205, 44, 36}
\definecolor{hidden-blue}{RGB}{194,232,247}
\definecolor{hidden-orange}{RGB}{243,202,120}
\definecolor{hidden-green}{RGB}{144,238,144}
\definecolor{hidden-pink}{RGB}{255,245,247}
\definecolor{hidden-black}{RGB}{20,68,106}

\tikzstyle{my-box}=[
    rectangle,
    draw=hidden-black,
    rounded corners,
    text opacity=1,
    minimum height=1.5em,
    minimum width=5em,
    inner sep=2pt,
    align=center,
    fill opacity=.5,
]
\tikzstyle{leaf}=[
    my-box, 
    minimum height=1.5em,
    fill=hidden-green!50, 
    text=black,
    align=left,
    font=\normalsize,
    inner xsep=2pt,
    inner ysep=10pt,
]

\tikzstyle{train_free}=[leaf, fill=trainfreeColor!50]
\tikzstyle{train}=[leaf, fill=trainColor!50]

\begin{figure*}[t!]
    \vspace{-2mm}
    \centering
    \resizebox{\textwidth}{!}{
        \begin{forest}
            forked edges,
            for tree={
                child anchor=west,
                parent anchor=east,
                grow'=east,
                anchor=west,
                base=left,
                font=\large,
                rectangle,
                draw=hidden-black,
                rounded corners,
                align=left,
                minimum width=4em,
                edge+={darkgray, line width=1pt},
                s sep=3pt,
                inner xsep=2pt,
                inner ysep=3pt,
                line width=0.8pt,
                ver/.style={rotate=90, child anchor=north, parent anchor=south, anchor=center},
            },
            where level=1{text width=9em,font=\normalsize,}{},
            where level=2{text width=15em,font=\normalsize,}{},
            where level=3{text width=14em,font=\normalsize,}{},
            where level=4{text width=10em,font=\normalsize,}{},
            [
                Concise and Adaptive Thinking, ver 
                [
                    ~Training-free~(\S\ref{sec:training_free}), train_free
                    [
                        ~Prompt-Guided~(\S\ref{sec:prompt_guided}), train_free
                        [
                            ~Concise CoT~\citep{Renze_2024}{,} 
                            Constraint CoT~\citep{nayab2025concisethoughtsimpactoutput}{,}  
                            TALE~\citep{han2025tokenbudgetawarellmreasoning}{,} \\
                            ~CoD~\citep{xu2025chaindraftthinkingfaster}{,}
                            ~Qwen3~\citep{yang2025qwen3technicalreport}{,}
                            ~NoThinking~\citep{ma2025reasoningmodelseffectivethinking}{,} \\
                            ~TokenComplexity~\citep{lee2025llmscompresschainofthoughttoken}{,}~CoUT~\citep{gong2025efficientreasoningchainunconscious}{,}{ etc.}
                            , train_free, text width=45.6em 
                        ]
                    ]
                    [
                        ~Pipeline~(\S\ref{sec:pipeline}), train_free
                        [
                            ~RouteLLM~\citep{ong2025routellmlearningroutellms}{,}~SoT~\citep{aytes2025sketchofthoughtefficientllmreasoning}{,}  
                            ThinkSwitcher~\citep{liang2025thinkswitcherthinkhardthink}{,} \\
                            ~SwitchCoT~\citep{zhang2025longshortcotinvestigating}{,}~SelfRoute~\citep{he2025selfrouteautomaticmodeswitching}{,}~CoThink~\citep{fan2025cothinktokenefficientreasoninginstruct}{,}\\~Think-to-Think~\citep{zhao2025t2adaptivetesttimescaling}{,}~SCoT~\citep{wang2025efficientreasoningllmsspeculative}{,}~Plan-and-Budget~\citep{lin2025planbudgeteffectiveefficient}{,} \\ ~CAR~\citep{lu2025prolongedreasoningneedcertaintybased}{,}
                            { etc.}
                            , train_free, text width=45.6em 
                        ]
                    ]
                    [
                        ~Decoding Manipulation~(\S\ref{sec:decoding_manipulation}), train_free
                        [
                            ~s1~\citep{muennighoff2025s1simpletesttimescaling}{,}~FlashThink~\citep{jiang2025flashthinkearlyexitmethod}{,}  
                            DEER~\citep{yang2025dynamicearlyexitreasoning}{,} \\
                            ~ConCISE~\citep{qiao2025conciseconfidenceguidedcompressionstepbystep}{,}~AlphaOne~\citep{zhang2025alphaonereasoningmodelsthinking}{,}~\citet{eisenstadt2025overclockingllmreasoningmonitoring}{,}\\
                            ~\citet{sheng2025reasoningstrengthplanninglarge}{,}
                            { etc.}
                            , train_free, text width=45.6em 
                        ]
                    ]
                    [
                        ~Model Merging~(\S\ref{sec:model_merging}), train_free
                        [
                            ~k1.5~\citep{kimiteam2025kimik15scalingreinforcement}{,} 
                            ~\citet{wu2025unlockingefficientlongtoshortllm}{,}  
                            ACM~\citep{yao2025activationguidedconsensusmerginglarge}{,}
                            { etc.}
                            , train_free, text width=45.6em 
                        ]
                    ]
                ]
                [
                    ~Training-based~(\S\ref{sec:training_based}), train
                    [
                        ~Fine-tunining Data Construction\\~(\S\ref{sec:data_construction}), train
                        [
                            ~ThoughtMani~\citep{liu2025thoughtmanipulationexternalthought}{,} 
                            Ada-R1~\citep{luo2025adar1hybridcotbileveladaptive}{,} 
                            ReCUT~\citep{jin2025recutbalancingreasoninglength}{,} \\
                            ~C3oT~\citep{kang2024c3otgeneratingshorterchainofthought}{,} 
                            A*-Thought~\citep{xu2025athoughtefficientreasoningbidirectional}{,} 
                            TokenSkip~\citep{xia2025tokenskipcontrollablechainofthoughtcompression}{,} \\ 
                            ~\citet{bercovich2025llamanemotronefficientreasoningmodels}{,} 
                            ~AutoL2S~\citep{luo2025autol2sautolongshortreasoning}{,}  
                            SBT~\citep{zhao2025letllmsbreakfree}{,} \\
                            ~Nemotron-CrossThink~\citep{akter2025nemotroncrossthinkscalingselflearningmath}{,} DAP~\citep{wu2025concisereasoningbiggains}{,} 
                            ConCISE~\citep{qiao2025conciseconfidenceguidedcompressionstepbystep}{,} \\
                            ~\citet{he2025selfrouteautomaticmodeswitching}{,} 
                            MoR~\citep{xiong2025mixturereasoningsteachlarge}{,} 
                            { etc.}
                            , train, text width=45.6em 
                        ]
                    ]
                    [
                        ~Variable Length Dataset~(\S\ref{sec:variable_length_dataset}), train
                        [
                            ~\citet{cai2025reasoningomnithoughtlargecot}{,} 
                            AM-DeepSeek-R1-Distilled~\citep{zhao202514millionopensourcedistilled}{,} 
                            ~\citet{zhang2025othinkr1intrinsicfastslowthinking}{,} \\
                            ~\citet{wen2025thinkpatterns21ksystematicstudyimpact}{,} 
                            DeepMath-103K~\citep{he2025deepmath103klargescalechallengingdecontaminated}{,} 
                            s1K-mix~\citep{muennighoff2025s1simpletesttimescaling}{,} \\
                            ~\citet{tian2025deepdistillenhancingllmreasoning}{,} 
                            { etc.}
                            , train, text width=45.6em 
                        ]
                    ]
                    [
                        ~Fine-tuning Approaches~(\S\ref{sec:finetune_approaches}), train
                        [
                            ~Long CoT Compression \\~Fine-tuning~, train
                            [
                                ~CoT-Valve\citep{ma2025cotvalvelengthcompressiblechainofthoughttuning}{,}
                                Thinker~\citep{chung2025thinkerlearningthinkfast}{,}\\
                                ~System-1.5 Reasoning~\citep{wang2025system15reasoningtraversallanguage}{,}
                                ~\citet{yu2025longshortchainofthoughtmixturesupervised}{,}\\
                                ~Adaptive GoGI-Skip~\citep{zhuang2025acceleratingchainofthoughtreasoninggoalgradient}{,} 
                                { etc.}
                                , train, text width=29.9em 
                            ]
                        ]
                        [
                            ~Short CoT Selection Fine-tuning~, train
                            [
                                ~\citet{munkhbat2025selftrainingelicitsconcisereasoning}{,}
                                ~\citet{yang2025thinkingoptimalscalingtesttimecompute}{,}\\
                                ~VeriThinker~\citep{chen2025verithinkerlearningverifymakes}{,} 
                                ~O1-Pruner~\citep{luo2025o1prunerlengthharmonizingfinetuningo1like}{,} 
                                { etc.}
                                , train, text width=29.9em 
                            ]
                        ]
                        [
                            ~Implicit CoT Fine-tuning~, train
                            [
                                ~\citet{pfau2024letsthinkdotdot}{,}
                                LightThinker~\citep{zhang2025lightthinkerthinkingstepbystepcompression}{,}\\
                                ~Heima~\citep{shen2025efficientreasoninghiddenthinking}{,}
                                ~CCoT~\citep{cheng2024compressedchainthoughtefficient}{,}\\
                                ~CoLaR~\citep{tan2025thinksilentlythinkfast}{,}
                                ~\citet{su2025tokenassortedmixinglatent}{,}
                                ~\citet{deng2024explicitcotimplicitcot}{,} \\
                                ~Coconut~\citep{hao2024traininglargelanguagemodels}{,}
                                ~\citet{yu2024distilling21}{,} 
                                ~\citet{liu2024expeditingelevatinglargelanguage}{,}\\
                                ~CODI~\citep{shen2025codicompressingchainofthoughtcontinuous}{,}
                                SoftCoT~\citep{xu2025softcotsoftchainofthoughtefficient}{,}\\
                                ~\citet{saunshi2025reasoninglatentthoughtspower}{,}
                                ~\citet{yu2025enhancingautoregressivechainofthoughtloopaligned}{,} 
                                { etc.}
                                , train, text width=29.9em 
                            ]
                        ]
                        [
                            ~DPO Variant Fine-tuning~, train
                            [
                                ~\citet{su2025underthinkingoverthinkingempiricalstudy}{,}
                                ~\citet{chen2025think23overthinkingo1like}{,}
                                ReCUT~\citep{jin2025recutbalancingreasoninglength}{,} \\
                                ~DTO~\citep{an2025dontthinklongerthink}{,}
                                TALE-PT~\citep{han2025tokenbudgetawarellmreasoning}{,} 
                                { etc.}
                                , train, text width=29.9em 
                            ]
                        ]
                        [
                            ~Other Fine-tuning~, train
                            [
                                ~Dualformer~\citep{su2025dualformercontrollablefastslow}{,}
                                ~\citet{wang2025adaptivedeepreasoningtriggering}{,}
                                ~\citet{li2025tldrlongreweightingefficient}{,} \\
                                ~\citet{zhang2025othinkr1intrinsicfastslowthinking}{,}
                                Qwen3~\citep{yang2025qwen3technicalreport}{,}
                                ~\citet{tencenthunyuanteam2025hunyuanturbosadvancinglargelanguage}{,} \\
                                ~\citet{wang2024guidinglanguagemodelreasoning}{,} 
                                ~TH2T~\citep{liu2025thinkthinkmitigatingoverthinking}{,}\\
                                ~MinD~\citep{zeng2025betterperfectunlockingefficient}{,}
                                QFFT~\citep{liu2025qfftquestionfreefinetuningadaptive}{,} 
                                { etc.}
                                , train, text width=29.9em 
                            ]
                        ]
                    ]
                    [
                        ~RL with Length Penalty~(\S\ref{sec:rl_with_length}), train
                        [
                            ~\citet{arora2025traininglanguagemodelsreason}{,} 
                            Light-R1~\citep{wen2025lightr1curriculumsftdpo}{,} 
                            ~\citet{yeo2025demystifyinglongchainofthoughtreasoning}{,} 
                            ~\citet{ling2025fasteasydeephard}{,} \\
                            ~LCPO\citep{aggarwal2025l1controllinglongreasoning}{,} 
                            Autothink~\citep{tu2025learningthinkshapingadaptive}{,} 
                            ~\citet{su2025thinkingfastrightbalancing}{,} \\
                            ~DAST\citep{shen2025dastdifficultyadaptiveslowthinkinglarge}{,}
                            ~\citet{song2025walkrunconcisellm}{,}
                            Short-RL\citep{yuan2025efficientrltrainingreasoning}{,}
                            ASRR\citep{zhang2025continuethinkingadaptivethinking}{,} \\
                            ~HAPO\citep{huang2025hapotraininglanguagemodels}{,}
                            ~\citet{wang2025adaptivethinkingmodepolicy}{,}
                            THINKPRUNE\citep{hou2025thinkprunepruninglongchainofthought}{,} \\
                            ~BRPO\citep{qi2025optimizinganytimereasoningbudget}{,}
                            ~REA-RL\citep{deng2025rearlreflectionawareonlinereinforcement}{,}
                            { etc.}
                            , train, text width=45.6em 
                        ]
                    ]
                    [
                        ~RL with GRPO-Variant~(\S\ref{sec:rl_with_grpo}), train
                        [
                            ~Ada-GRPO~\citep{wu2025armadaptivereasoningmodel}{,} 
                            DeGRPO~\citep{fang2025thinklessllmlearnsthink}{,}  
                            SelfBudgeter~\citep{li2025selfbudgeteradaptivetokenallocation}{,} \\
                            ~DIET~\citep{chen2025overthinkersdietcuttingtoken}{,}
                            Elastic Reasoning~\citep{xu2025scalablechainthoughtselastic}{,}
                            HGPO~\citep{jiang2025thinkneedlargehybridreasoning}{,} \\
                            ~AdaCtrl~\citep{huang2025adactrladaptivecontrollablereasoning}{,}
                            ACPO~\citep{cheng2025incentivizingdualprocessthinking}{,}
                            ConciseR~\citep{song2025walkrunconcisellm}{,} \\
                            ~LC-R1~\citep{cheng2025optimizinglengthcompressionlarge}{,} 
                            { etc.}
                            , train, text width=45.6em 
                        ]
                    ]
                    [
                        ~RL with Difficulty-awareness\\~(\S\ref{sec:rl_with_difficulty}), train
                        [
                            ~DAST~\citep{shen2025dastdifficultyadaptiveslowthinkinglarge}{,} 
                            SelfBudgeter~\citep{li2025selfbudgeteradaptivetokenallocation}{,}  
                            DIET~\citep{chen2025overthinkersdietcuttingtoken}{,} \\
                            ~AdaCtrl~\citep{huang2025adactrladaptivecontrollablereasoning}{,}
                            ALP~\citep{xiang2025justthinkingefficientreasoning}{,}
                            ACPO~\citep{cheng2025incentivizingdualprocessthinking}{,} \\
                            ~\citet{ling2025fasteasydeephard}{,}
                            { etc.}
                            , train, text width=45.6em 
                        ]
                    ]
                    [
                        ~RL with Thinking Mode~(\S\ref{sec:rl_with_think_modes}), train
                        [
                            ~AdaptThink~\citep{zhang2025adaptthinkreasoningmodelslearn}{,} 
                            Autothink~\citep{tu2025learningthinkshapingadaptive}{,}  
                            ~\citet{lou2025adacotparetooptimaladaptivechainofthought}{,} 
                            ~\citet{jiang2025thinkneedlargehybridreasoning}{,}
                            { etc.}
                            , train, text width=45.6em 
                        ]
                    ]
                    [
                        ~Other RL~(\S\ref{sec:rl_other}), train
                        [
                            ~L2T~\citep{wang2025learningthinkinformationtheoreticreinforcement}{,} 
                            TWYN~\citep{yang2025thinkneedselfadaptivechainofthought}{,}  
                            ConciseRL~\citep{dumitru2025conciserlconcisenessguidedreinforcementlearning}{,} \\
                            ~\citet{xie2025interleavedreasoninglargelanguage}{,}
                            ~\citet{yi2025shorterbetterguidingreasoningmodels}{,}
                            ~\citet{fatemi2025concisereasoningreinforcementlearning}{,}
                            REO-RL~\citep{gao2025faroptimalreasoningefficiency}{,} \\
                            ~Long\(\otimes\)Short~\citep{ning2025thoughtsgeneratedequalefficient}{,}
                            { etc.}
                            , train, text width=45.6em 
                        ]
                    ]
                ]
            ]
        \end{forest}
    }
    \caption{Taxonomy of Concise and Adaptive Thinking in LRMs.}
    \label{fig:taxonomy}
\end{figure*}
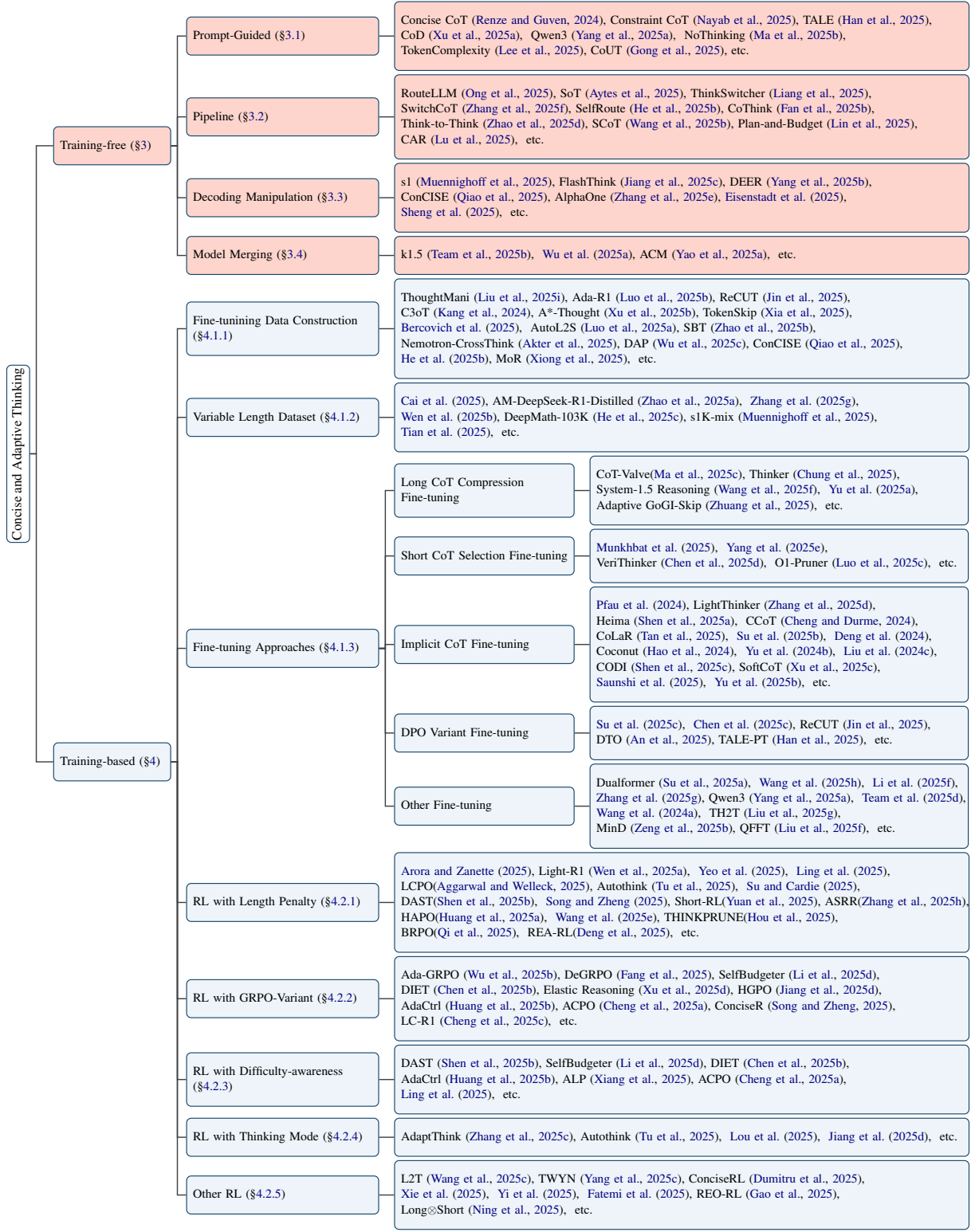

\section{Evaluation}
\subsection{Models}
There have been a substantial number of LRMs in the community, including open-source weights and commercial APIs. Many of them serve as the base model in the benchmark experiments through concise or adaptive thinking approaches. Among these LRMs, distill/SFT based LRM - DeepSeek-R1-Distill series (1.5B / 7B) and RL based LRM - QwQ~\cite{qwq32b} exhibit the highest adoption rates. Some studies~\cite{zhu2025largereasoningmodelssave,liu2025thoughtmanipulationexternalthought} show that RL based LRMs frequently engage in reasoning even when explicitly instructed to skip directly to the final answer, while SFT based LRMs consistently follow concise thinking instructions. This discrepancy may lead to the situation where some methods are not universally applicable across all LRMs. Table~\ref{table:basemodel_dataset_train_free} and Table~\ref{table:basemodel_dataset_train} summarize the base models used in selected concise and adaptive thinking methods.

\newcolumntype{L}[1]{>{\raggedright\arraybackslash}p{#1}} 
\newcolumntype{C}[1]{>{\centering\arraybackslash}p{#1}}   
\newcolumntype{R}[1]{>{\raggedleft\arraybackslash}p{#1}}  

\begin{table*}[!ht]
\centering
\resizebox{1\textwidth}{!}{
\begin{tabular}{@{}ccc@{}}
\hline
\multicolumn{1}{c}{\textbf{Methods}}  & \multicolumn{1}{c}{\textbf{Base Model}}& \multicolumn{1}{c}{\textbf{Datasets}} \\ \hline
CCoT~\cite{Renze_2024} & GPT3.5, GPT4 &  - \\ \hline
Constrained-CoT~\cite{nayab2025concisethoughtsimpactoutput} & Falcon-40b-instruct, Llama2-70b-chat-hf  &   GSM8K, SVAMP~\cite{patel2021nlpmodelsreallyable}, ASDIV~\cite{miao2021diversecorpusevaluatingdeveloping} \\ \hline
TALE~\cite{han2025tokenbudgetawarellmreasoning} & \makecell[c]{GPT-4o, GPT-4o-mini, o3-mini,\\ Yi-lightning, Lllama3.1-8B-Instruct}  &  GSM8K, GSM8K-Zero, MathBench~\cite{liu2024mathbenchevaluatingtheoryapplication} \\ \hline
CoD~\cite{xu2025chaindraftthinkingfaster} & GPT-4o, Claude 3.5 Sonnet &  GSM8K, BBH~\cite{suzgun2022challengingbigbenchtaskschainofthought} \\ \hline
NoThinking~\cite{ma2025reasoningmodelseffectivethinking} & R1-Distill-Qwen-32B & \makecell[c]{AIME24/25, AMC23,\\ OlympiadBench~\cite{he2024olympiadbenchchallengingbenchmarkpromoting}, LiveCodeBench, \\ MiniF2F~\cite{zheng2022minif2fcrosssystembenchmarkformal}, ProofNet~\cite{azerbayev2023proofnetautoformalizingformallyproving}} \\ \hline
CoUT~\cite{gong2025efficientreasoningchainunconscious} & \makecell[c]{GPT-4o, Claude 3.5 Sonnet,\\o3-mini, QWQ} & \makecell[c]{GSM8K, SVAMP,\\ MathQA~\cite{amini2019mathqainterpretablemathword}, etc.} \\ \hline
TokenComplexity~\cite{lee2025llmscompresschainofthoughttoken} & \makecell[c]{GPT-4o, GPT-4o mini,\\Claude 3.5 Sonnet/Haiku, Llama3.3-70B} & GSM8K, MATH-500, MMLU-Pro Math~\cite{wang2024mmluprorobustchallengingmultitask} \\ \hline
RouteLLM~\cite{ong2025routellmlearningroutellms} & Claude 3 Opus/Sonnet, Llama3.1-8B/70B & \makecell[c]{GSM8K, MMLU~\cite{hendrycks2021measuringmassivemultitasklanguage},\\MT-Bench~\cite{zheng2023judgingllmasajudgemtbenchchatbot}}  \\ \hline
SoT~\cite{aytes2025sketchofthoughtefficientllmreasoning} & \makecell[c]{GPT-4o, Claude 3.5 Sonnet,\\Qwen2.5 7B/14B/32B,\\ Llama3.1 8B, Llama3.2 11B}  & \makecell[c]{GSM8K, DROP~\cite{dua2019dropreadingcomprehensionbenchmark},\\ CommonsenseQA~\cite{talmor2019commonsenseqaquestionansweringchallenge}, \\OpenBookQA~\cite{mihaylov2018suitarmorconductelectricity}, LogiQA~\cite{liu2020logiqachallengedatasetmachine}, \\StrategyQA~\cite{geva2021didaristotleuselaptop}, HotpotQA~\cite{yang2018hotpotqadatasetdiverseexplainable}, etc.} \\ \hline
ThinkSwither~\cite{liang2025thinkswitcherthinkhardthink} &  R1-Distill-Qwen-1.5B/7B/14B & \makecell[c]{GSM8K, MATH-500, AIME24/25, \\LiveAoPS~\cite{mahdavi2025leveragingonlineolympiadlevelmath}, OmniMATH~\cite{gao2024omnimathuniversalolympiadlevel},\\OlympiadBench} \\ \hline
SelfRoute~\cite{he2025selfrouteautomaticmodeswitching} &  R1-Distill-Qwen-7B/32B & \makecell[c]{GSM8K, Math-500, AIME24,\\ GPQA~\cite{rein2023gpqagraduatelevelgoogleproofqa}, ARC-C~\cite{clark2018thinksolvedquestionanswering}} \\ \hline
CoThink~\cite{fan2025cothinktokenefficientreasoninginstruct} &  \makecell[c]{DAPO-Qwen-32B,\\ R1-Distill-Qwen-32B, QwQ} & GSM8K, MATH-500, AIME24 \\ \hline
SCoT~\cite{wang2025efficientreasoningllmsspeculative} &  \makecell[c]{R1-Distill-Qwen-32B,\\R1-Distill-Llama-70B} & \makecell[c]{GSM8K, MATH, GaoKao~\cite{liao2024mariomathreasoningcode},\\ CollegeMath~\cite{tang2024mathscalescalinginstructiontuning}, OlympiadBench}  \\ \hline
Plan and Budget~\cite{lin2025planbudgeteffectiveefficient} &  \makecell[c]{R1-Distill-Qwen-32B, R1-Distill-Llama-70B,\\QwQ-32B, o4-mini} & \makecell[c]{MATH-500,\\ NaturalInstructions~\cite{mishra2022crosstaskgeneralizationnaturallanguage},\\ TravelPlanner~\cite{xie2024travelplannerbenchmarkrealworldplanning}}  \\ \hline
CAR~\cite{lu2025prolongedreasoningneedcertaintybased} &  Qwen2.5-7B-Instruct, Llama3.1-8B-Instruct & GSM8K, MathQA, StrategyQA  \\ \hline
s1~\cite{muennighoff2025s1simpletesttimescaling} &  \makecell[c]{o1/o1-preview/o1-mini,\\DeepSeek R1, QwQ, etc.} & MATH-500, AIME24, GPQA  \\ \hline
DEER~\cite{yang2025dynamicearlyexitreasoning} &  \makecell[c]{R1-DistillQwen-7B/14B/32B,\\ Qwen3-14B, QwQ} & \makecell[c]{GSM8K, MATH-500, AIME24, AMC23, GPQA, \\ HumanEval~\cite{chen2021evaluatinglargelanguagemodels}, BigCodeBench~\cite{zhuo2025bigcodebenchbenchmarkingcodegeneration},\\LiveCodeBench}  \\ \hline
ThinkOrNot~\cite{yong2025thinknotexploringthinking} &  R1-DistillQwen-32B, QwQ & GSM8K, AIME25  \\ \hline
ConCISE~\cite{qiao2025conciseconfidenceguidedcompressionstepbystep} &  R1-DistillQwen-1.5B/7B & GSM8K, Math-500, AIME24, GPQA  \\ \hline
$short\text{-}m@k$~\cite{hassid2025dontoverthinkitpreferring} &  \makecell[c]{Llama-3.3-Nemotron-Super49B-v1,\\ R1-DistillQwen-32B, QwQ} & AIME24/25, HMMT~\cite{balunović2025matharenaevaluatingllmsuncontaminated}  \\ \hline
\makecell[c]{Unlocking~\cite{wu2025unlockingefficientlongtoshortllm},\\ ACM~\cite{yao2025activationguidedconsensusmerginglarge}} &  \makecell[c]{Qwen2.5-Math-7B,\\R1-Distill-Qwen-7B} & \makecell[c]{GSM8K, MATH-500, AIME24, \\Minerva Math~\cite{lewkowycz2022solvingquantitativereasoningproblems},\\Olympiadbench, CollegeMath}\\ \hline
\end{tabular}}
\caption{Summary of benchmark datasets and base models used in selected training-free concise and adaptive thinking methods.}
\label{table:basemodel_dataset_train_free}
\end{table*}

\begin{table*}[!ht]
\centering
\resizebox{1\textwidth}{!}{
\begin{tabular}{@{}ccc@{}}
\hline
\multicolumn{1}{c}{\textbf{Methods}}  & \multicolumn{1}{c}{\textbf{Base Model}}& \multicolumn{1}{c}{\textbf{Datasets}} \\ \hline
ThoughtMani\cite{liu2025thoughtmanipulationexternalthought}& \makecell[c]{Qwen-Max/Plus, Qwen2.5-7B-Instruct, \\Qwen2.5-3B-Instruct, QwQ-32B} & \makecell[c]{GSM8k, MATH-500, AIME24, \\LiveBench~\cite{white2025livebenchchallengingcontaminationlimitedllm}, WildJailbreak~\cite{jiang2024wildteamingscaleinthewildjailbreaks}}  \\ \hline
Ada-R1\cite{luo2025adar1hybridcotbileveladaptive}& \makecell[c]{DeepSeek-R1-DistillQwen-7B, \\DeepSeek-R1-Distill-Qwen-1.5B} & GSM8K, MATH, AIME24  \\ \hline
C3oT~\cite{luo2025adar1hybridcotbileveladaptive}& LLaMA2-Chat-7B/13B & GSM8K, MathQA, ECQA~\cite{aggarwal-etal-2021-explanations}, StrategyQA  \\ \hline
TokenSkip~\cite{xia2025tokenskipcontrollablechainofthoughtcompression}& LLaMA-3.1-8B-Instruct, Qwen2.5-Instruct series & GSM8K, MATH \\ \hline
Llama-Nemotron~\cite{bercovich2025llamanemotronefficientreasoningmodels}& LN-Nano/Super/Ultra & MATH-500, AIME24/25, GPQA, LiveCodeBench \\ \hline
SBT~\cite{zhao2025letllmsbreakfree}& \makecell[c]{Qwen2.5-Math-1.5B/7BInstruct, \\Llama-3.2-1B, Llama-3.1-8B-Instruct} & GSM8K, MATH-500, AIME, AMC \\ \hline
A*Thought~\cite{xu2025athoughtefficientreasoningbidirectional}& \makecell[c]{QwQ-32B, \(s1.1-32B\), \\ DeepSeek-Rl-Distill-Qwen \(32B\)} & GSM8K, MATH-500, AMC23, OlympiadBench \\ \hline
Nemotron-CrossThink~\cite{akter2025nemotroncrossthinkscalingselflearningmath}& Qwen2.5-7B, Qwen2.5-32B & \makecell[c]{MATH-500, AMC23, MMLU, MMLU-Pro \\ AgiEval, GPQA, SuperGPQA~\cite{pteam2025supergpqascalingllmevaluation}} \\ \hline
AutoL2S~\cite{luo2025autol2sautolongshortreasoning}& Qwen2.5-3B-Instruct, Qwen2.5-7B-Instruct & GSM8K, MATH-500, GPQA, OlympiadBench \\ \hline
DAP~\cite{wu2025concisereasoningbiggains}& Qwen2.5 series, Llama3.1 series & \makecell[c]{GSM8K, MATH-500, AIME24/25, OlympiadBench,\\ MinervaMath, GPQA, GaoKao2023, MMLU-STEM} \\ \hline
~\citet{chen2025think23overthinkingo1like}& Qwen-QwQ-32B-Preview, DeepSeek-R1 & GSM8K, MATH-500, ASDIV  \\ \hline
LS-Mixture SFT~\cite{yu2025longshortchainofthoughtmixturesupervised}& Qwen2.5-32B-Instruct & Math-500, AIME24, GPQA \\ \hline
DTO~\cite{an2025dontthinklongerthink}& \makecell[c]{DeepSeek-R1-Distill-Qwen-1.5B, \\ DeepScaleR-1.5B-Preview} & GSM8K, MATH, AIME24/25, AMC23, Gaokao \\ \hline 
~\citet{deng2024explicitcotimplicitcot} & GPT-2 Small, Mistral 7B, Phi-3 3.8B & Multi-digit multiplication, Grade school math  \\ \hline
~\citet{yu2024distilling21} & Llama-2-70B-chat & OASST2, MTbench  \\ \hline
CODI~\cite{shen2025codicompressingchainofthoughtcontinuous}& GPT-2, LLaMA3.2-1b-Instruct & SVAMP, GSM-
HARD, MultiArith \\ \hline 
~\citet{arora2025traininglanguagemodelsreason}& R1-Distill-Qwen-1.5B/7B & GSM8K, MATH-500, AIME24  \\ \hline
Demystifying~\cite{yeo2025demystifyinglongchainofthoughtreasoning}& Llama3.1-8B & MATH-500, AIME24, TheoremQA, MMLU-pro-1k  \\ \hline
L1~\cite{aggarwal2025l1controllinglongreasoning}&  R1-Distill-Qwen-1.5B & AIME25, MATH, AMC, OlympiadBench, GPQA, LSAT, MMLU  \\ \hline
AutoThink~\cite{tu2025learningthinkshapingadaptive}& \makecell[c]{R1-Distill-Qwen-1.5B/7B,\\DeepScaleR-Preview-1.5B} & MATH, AIME24, AMC23, Minerva, Olympiad  \\ \hline
AdaCoT~\cite{lou2025adacotparetooptimaladaptivechainofthought}& Internal 15B/150B MoE   & \makecell[c]{MATH, AIME24/25, GPQA, LiveCodeBench, \\SuperGPQA, OlympiadBench, MMLU Pro, \\Chinese SimpleQA~\cite{he2024chinesesimpleqachinesefactuality}, SimpleQA~\cite{wei2024measuringshortformfactualitylarge},\\ LiveBench, Swe-Bench~\cite{jimenez2024swebenchlanguagemodelsresolve}, \\SysBench~\cite{qin2024sysbenchlargelanguagemodels}, Kor-Bench~\cite{ma2025korbenchbenchmarkinglanguagemodels}, etc.}  \\ \hline
DAST~\cite{shen2025dastdifficultyadaptiveslowthinkinglarge}& R1-Distill-Qwen-7B/32B   & MATH-500, AIME24, GPQA  \\ \hline
ConciseR~\cite{song2025walkrunconcisellm}& Qwen2.5-Math-7B   & MATH-500, AIME24, AMC23, Minerva, OlympiadBench  \\ \hline
Short-RL~\cite{yuan2025efficientrltrainingreasoning}& \makecell[c]{Qwen2.5-7B, DeepScaleR,\\Open-Reasoner-Zero, SimpleRL-Reason} & \makecell[c]{MATH-500, AIME24, AMC23, \\Minerva Math, OlympiadBench} \\ \hline
ASRR~\cite{zhang2025continuethinkingadaptivethinking}& R1-Distill-Qwen-1.5B/7B & MATH-500, AIME24, AMC23, OlympiadBench, GSM8K \\ \hline
HAPO~\cite{huang2025hapotraininglanguagemodels}& \makecell[c]{R1-Distill-Qwen-1.5B,\\DeepScaleR-1.5B-Preview,\\Qwen-2.5-1.5B-Instruct} & GSM8K, MATH-500, AIME24 \\ \hline
AdaCtrl~\cite{huang2025adactrladaptivecontrollablereasoning}&  Qwen2.5-7B-Instruct & GSM8K, MATH-500, AIME24/25 \\ \hline
ThinkPrune~\cite{hou2025thinkprunepruninglongchainofthought}&  \makecell[c]{R1-Distill-Qwen-1.5B,\\ DeepScaleR1.5B-Preview, QwQ} & MATH-500, AIME24, AMC23, OlympiadBench \\ \hline
BRPO~\cite{qi2025optimizinganytimereasoningbudget}&  R1-Distill-Qwen-1.5B & AIME24, AMC22, MATH-500, Minerva Math, OlympiadBench  \\ \hline
REA-RL~\cite{deng2025rearlreflectionawareonlinereinforcement}&  R1-Distill-Qwen-7B & \makecell[c]{GSM8K, Math-500, AIME24,\\AMC23, Gaokao23, OlympiadBench}  \\ \hline
ARM~\cite{wu2025armadaptivereasoningmodel}&  Qwen2.5-Base-3B/7B/14B & \makecell[c]{GSM8K, MATH, AIME25, SVAMP,\\CommonsenseQA, OpenBookQA, BBH}  \\ \hline
DeGRPO~\cite{fang2025thinklessllmlearnsthink}&  R1-Distill-Qwen-1.5B &  GSM8K, MATH-50, AIME24, Minerva Algebra, 0  \\ \hline
SelfBudgeter~\cite{li2025selfbudgeteradaptivetokenallocation}& R1-Distill-Qwen-1.5B & GSM8K, MATH  \\ \hline
DIET~\cite{chen2025overthinkersdietcuttingtoken}& R1-Distill-Qwen-1.5B & MATH-500, AIME24, AMC23, OlympiadBench, Minerva \\ \hline
Elastic Reasoning~\cite{xu2025scalablechainthoughtselastic}& \makecell[c]{DeepScaleR-1.5B-Preview,\\ DeepCoder-14B-Preview} & \makecell[c]{MATH-500, AIME24, AMC,\\ OlympiadBench, Minerva Math, LiveCodeBench,\\ Codeforces, HumanEval+~\cite{liu2023codegeneratedchatgptreally}} \\ \hline
HGPO~\cite{jiang2025thinkneedlargehybridreasoning}&  Qwen-2.5-math-base 1.5B/7B & \makecell[c]{MATH-500, AIME24, AMC23, OlympaidBench,\\ LiveCodeBench, MBPP~\cite{austin2021programsynthesislargelanguage},\\ MBPP+~\cite{liu2024evaluatinglanguagemodelsefficient}, \\AlpacaEval 2.0~\cite{dubois2025lengthcontrolledalpacaevalsimpleway}, ArenaHard~\cite{li2024crowdsourceddatahighqualitybenchmarks}}  \\ \hline
AdaptThink~\cite{zhang2025adaptthinkreasoningmodelslearn}&   R1-Distill-Qwen-1.5B/7B &  GSM8K, MATH-500, AIME24  \\ \hline
L2T~\cite{wang2025learningthinkinformationtheoreticreinforcement}&  \makecell[c]{R1-Distill-Qwen-1.5B,\\ DeepScaleR-1.5B-Preview} & MATH-500, AIME24/25, AMC23, Minerva Math  \\ \hline
TWYN~\cite{yang2025thinkneedselfadaptivechainofthought}&   R1-Distill-Qwen-1.5B/7B & MATH-500, AIME24, AMC23, Minerva Math, OlympiadBench  \\ \hline
\end{tabular}}
\caption{Summary of benchmark datasets and base models used in selected training based concise and adaptive thinking methods.}
\label{table:basemodel_dataset_train}
\end{table*}

\subsection{Datasets}
Since LRMs are widely trained and evaluated on reasoning tasks, they are more likely to generate lengthy responses in these domains. In order to evaluate the reasoning efficiency, mathematics and coding datasets are commonly used, including: GSM8K~\cite{cobbe2021trainingverifierssolvemath}, MATH-500~\cite{hendrycks2021measuringmathematicalproblemsolving}, AIME(24 \& 25), AMC2023, LiveCodeBench~\cite{jain2024livecodebenchholisticcontaminationfree}, etc. Sys2Bench~\cite{parashar2025inferencetimecomputationsllmreasoning} is a comprehensive benchmark designed for complex reasoning tasks. It contains 11 datasets across 5 categories like Algorithmic Reasoning, Planning, Logical Reasoning, etc. Table~\ref{table:basemodel_dataset_train_free} and Table~\ref{table:basemodel_dataset_train} summarize the benchmark datasets used in selected concise and adaptive thinking methods.


Beyond datasets measuring performance of LRMs, other benchmarks assess the reasoning efficiency or the extent of overthinking of LRMs. DNR Bench~\cite{hashemi2025dnrbenchbenchmarkingoverreasoning} is designed with 150 adversarial samples across 5 categories to expose the overthinking problem of LRMs. \citet{li2025thinkbenchevaluatingthinkingefficiency} introduce Think-Bench to comprehensively evaluate the reasoning efficiency and accuracy of LRMs. It contains 1,375 samples from mathematics, physics, and chemistry. Claude 3.7 is used to generate reasoning chains. Based on the full reasoning paths, they identify critical steps which cannot be omitted. ICBench~\cite{yang2025longtoshortfreelunchinvestigating} is another benchmark which is used to measure inconsistency in LRMs before and after adaptive thinking optimization across three dimensions. Therefore, it is possible to observe whether the adaptive thinking approaches have introduced any potential risks in addition to the degradation of accuracy.

\subsection{Metrics}
As mentioned above, the fundamental objective of adaptive thinking lies in achieving the balance between reasoning efficiency and performance, specifically, maintaining reasoning accuracy while reducing reasoning length. From this perspective, two primary metrics emerge to evaluate adaptive thinking methods: \textit{accuracy} and \textit{number of inference tokens}. Beyond these two metrics, recent researches have introduced various metrics to quantify reasoning efficiency from different aspects. This section presents a comprehensive review of these emerging evaluation paradigms.

\noindent\textbf{Accuracy}, the fraction of questions solved correctly, is widely used to measure the performance on a dataset. The hyper-parameters of generation are specified according to the LRMs.

\noindent\textbf{pass@k} measures the probability of obtaining at least one correct output among $k$ randomly selected samples out of $n$ generated completions per problem. Note that \textit{accuracy} is equivalent to $pass@1$. $pass@k$ is widely adopted by those parallel scaling approaches for efficient reasoning~\cite{ma2025reasoningmodelseffectivethinking,zhang2025reasoningmodelsknowtheyre}.

\noindent\textbf{Number of Inference Tokens} is the average number of output tokens produced by the LLMs in their response across questions in the dataset.

\noindent\textbf{Latency} measures the time (e.g., seconds) required to generate a final response to end users. This is the essential goal in real-world applications. However, latency exhibits strong dependence on hardware specifications, inference framework, and model size, etc. Consequently, most adaptive thinking methods exclude this metric from their evaluation results.

\noindent\textbf{Speed up Ratio} is a derived metric from \textit{Number of Inference Tokens} and \textit{Latency}. It quantifies the efficiency gap of LRMs before and after adaptive thinking optimization, for example, by dividing the number of inference tokens before optimization by that after optimization.

\noindent\textbf{Reasoning Verbosity} is proposed by \citet{cai2025reasoningomnithoughtlargecot} to score problem-CoT pairs to train length-optimal LRMs. The lower verbosity means straightforward expression with little or no elaboration, while the higher verbosity means a deeper and exhaustive exploration. The raw \textit{Reasoning Verbosity} from LLM judge is fused with the normalized length score to get the final \textit{Reasoning Verbosity} score.

\noindent\textbf{CoT Trigger Rate} is used to measure the fraction of queries that trigger long CoT (i.e \textit{<think>}) when the LRM is applied to production traffic, which reflects unbiased user query distributions. \citet{lou2025adacotparetooptimaladaptivechainofthought} use this metric to demonstrate the significant reduction in CoT usage, which directly leads to the saving of computational resources.

\noindent\textbf{Outcome/Process Efficiency} are proposed by~\citet{chen2025think23overthinkingo1like} to assess the efficiency of LRMs. \textit{Outcome Efficiency} is based on the observation that the initial round of solutions provides the correct answer in most cases, while the subsequent solutions, either alternative or self-reflection solutions, contribute minimally to the accuracy. So the \textit{Outcome Efficiency} is calculated as the fraction of tokens that first arrive at the correct answer. Intuitively, when a LRM tends not to generate redundant verification after producing the correct answer, this indicates higher \textit{Outcome Efficiency}. While \textit{Outcome Efficiency} quantifies the contribution of tokens in the response to the final accuracy from a results-oriented perspective, \textit{Process Efficiency} assesses the diversity of solutions/steps in the reasoning chains from a procedural point. If a CoT response contains multiple reflection and verification steps but exhibits redundancy among these steps, it indicates low \textit{Process Efficiency} in the model.

\noindent\textbf{HCA/SCA/CCA}, proposed by ~\citet{nayab2025concisethoughtsimpactoutput}, are three metrics to evaluate the capability of an LLM to provide correct and concise responses. \textit{HCA}, or HCA($k$), Hard-$k$ Concise Accuracy, means the fraction of correct outputs that do not exceed a user-specified length $k$. \textit{SCA}, or SCA($k$,$\alpha$), Soft-$k$ Concise Accuracy, differs from \textit{HCA($k$)} in its penalty mechanism: instead of assigning a zero score to those responses with correct answers exceeding the user-specified length limit, it imposes an exponential decay-based penalty on responses that violate the length constraint. \textit{CCA}, or CCA($k$,$\alpha$,$\beta$), Consistent Concise Accuracy, further takes the variation of the lengths among all the outputs into consideration. This metric encourages length consistency of responses as that the predictability of inference time is very important in real-world applications.

\noindent\textbf{\underline{E}fficiency-Aware \underline{E}ffectiveness \underline{E}valuation}, or $\mathcal{E}^3$~\cite{lin2025planbudgeteffectiveefficient} jointly captures the accuracy and reasoning length, using a simple formula $\mathcal{E}^3=\frac{A^2}{T}$ where $A$ is the average accuracy across a dataset and $T$ denotes the average reasoning length. The square of accuracy aims to emphasize the correctness rather than indiscriminately reducing length, which would compromise accuracy.

\noindent\textbf{InfoBias}~\cite{yong2025thinknotexploringthinking} is introduced to capture how closely the reasoning trajectory aligns with the ground truth using mutual information. \textbf{InfoGain} quantifies the contribution of each reasoning step by measuring how much uncertainty is reduced by incorporating this step.

\noindent\textbf{Overthinking Score}~\cite{cuadron2025dangeroverthinkingexaminingreasoningaction} is a LLM score ranging from 0 to 10 that analyzes model trajectories from three key patterns: Analysis Paralysis, Rogue Actions, and Premature Disengagement. A higher score indicates a more severe overthinking behavior. 

\noindent\textbf{Think-Bench}~\cite{li2025thinkbenchevaluatingthinkingefficiency} introduce six complementary metrics to evaluate the reasoning efficiency from token usage, inference dynamics, and reflective quality perspectives, including \textit{Tokens}, \textit{First Correct Tokens}, \textit{Efficiency}, \textit{Reflection Quality}, \textit{Reflection Tokens}, \textit{Thought Num}.

\noindent\textbf{Think Density} is proposed by~\citet{li2025dynamicmindtrimodethinkinglarge} to measure the trade-off between accuracy and number of used tokens. Given a prompt $q$ and its $k$ generated responses, the \textit{Think Density} is calculated as the average accuracy of $k$ responses divided by the average number of tokens.

\subsection{Existing Problems}
Current evaluations of adaptive thinking methods in LRMs face two primary challenges. Firstly, the majority of these methods are exclusively evaluated on reasoning-focused benchmarks, leaving their impacts on models' fundamental capabilities underexplored (as shown in Table~\ref{table:basemodel_dataset_train_free} and Table~\ref{table:basemodel_dataset_train}). Secondly, there are no unified evaluation standards, including base model, datasets, metrics, and hyper-parameters of inference. These discrepancies significantly hinder the standardization and comparative analysis of related research efforts.

\section{Training-free Methods}
\label{sec:training_free}
Training-free methods, owing to their convenience, are widely explored to rapidly address the issue of excessive reasoning length in LRMs. We categorize these approaches into three classes: (1) \textbf{prompt-guided strategies}, which simply leverage the instruction-following capability of LRMs by employing carefully crafted prompts to steer the model's reasoning process according to predefined requirements. This includes enforcing constraints such as simplified reasoning steps (e.g., ``\textit{Be concise}'') or systematic limitations on reasoning depth and token usage. (2) \textbf{Pipeline based methods}, which design modular workflows to strategically offload computational overhead from LRMs. (3) \textbf{Decoding manipulation}, which focuses on strategically intervening in LRMs' reasoning processes to enable models to break free from protracted reasoning sequences. (4) \textbf{Model merging}, which consolidates a slow-thinking LRM and a fast-thinking LLM to a single one, and this single model is expected to balance fast-slow thinking and thus achieve adaptive thinking.

\subsection{Prompt-Guided}
\label{sec:prompt_guided}
These works aim to force LRMs switch between long CoT and short CoT through prompt engineering, like adding ``\textit{Be concise}''~\cite{Renze_2024} or ``\textit{limit the answer length to xx words.}''~\cite{nayab2025concisethoughtsimpactoutput} to the original prompt.

\subsubsection{Token Budget}
Responses of similar lengths vary greatly in correctness~\cite{jiang2025makesgoodreasoningchain}. TALE~\cite{han2025tokenbudgetawarellmreasoning} further introduces the concept of \textbf{token budget}. By searching for an optimal token budget for a given problem, the LRMs can be prompted to reason under the token budget, achieving a satisfying trade-off between efficiency and performance.

\subsubsection{Thinking Mode}
Instead of setting an overall token budget in prompt, CoD~\cite{xu2025chaindraftthinkingfaster} prompts the model to be concise for each reasoning step (e.g., 5 words at most). They argue that this prompt strategy is aligned with human cognitive processes when solving complex tasks. Inspired by unconscious thought theory, \citet{gong2025efficientreasoningchainunconscious} propose CoUT which first encourages LRMs to think internally and then efficiently generates the answer with several well-crafted token efficient strategies. \citet{lee2025llmscompresschainofthoughttoken} conduct a systematic study of the relationship between reasoning length and model performance across a diverse range of prompts and reveal that each task has a minimal number of tokens required for successful problem-solving, which they call it as ``token complexity''.

\subsubsection{No Thinking}
An alternative methodology employs a more blunt strategy that completely disables the explicit thinking process of LRMs. For example, ~\citet{yang2025qwen3technicalreport} force the thinking box to be empty (``\textit{<think>\textbackslash n\textbackslash n</think>}''), while NoThinking~\cite{ma2025reasoningmodelseffectivethinking} simply bypasses the thinking process with prompt like ``\textit{<think>Okay, I have finished thinking.</think>}''. They find that such a NoThinking strategy can achieve comparable performance ($pass@k$) with default Thinking setting while using far fewer tokens. \citet{zhu2025largereasoningmodelssave} find that LRMs do not always follow the instruction to skip thinking under NoThinking setting. They investigate three thinking modes: (1) No thinking, which directly generates the answer without thinking. (2) Explicit thinking, which reinvigorates the thinking process with an additional ``</think>'' token. (3) Implicit thinking, which re-engages the thinking process but no additional ``</think>'' token appended. The internal states and activation patterns are analyzed in different thinking modes, so that they can be used to guide future research on the reliability of reasoning.

Although prompt-guided methods can achieve the compression of reasoning chain in a very simple way, they essentially rely on the model's instruction following ability on the constraints and criteria in the prompt. On the one hand, some studies~\cite{zhu2025largereasoningmodelssave} have shown that LRMs do not necessarily follow the prompt for output. For example, they may frequently engage in reasoning even when explicitly instructed to skip directly to the final answer. On the other hand, \citet{yang2025longtoshortfreelunchinvestigating} indicate that strategies like \textit{Token Budget} or \textit{NoThinking} are not free lunches. They may greatly increase the inconsistency of the model before and after optimization, thus hiding key decision factors in the model and introducing potential risks in addition to the degradation of accuracy.

\subsection{Pipeline}
\label{sec:pipeline}
In contrast to prompt-guided approaches, which directly leverage the intrinsic instruction-following capabilities of LRMs to enable concise and adaptive reasoning, pipeline-based methods aim to design modular workflows to strategically offload computational overhead from LRMs.

\subsubsection{Router}
Among these approaches, router-based architectures~\cite{pan2024dynathinkfastslowdynamic,yao2024hdflowenhancingllmcomplex,ong2025routellmlearningroutellms,ding2024hybridllmcostefficientqualityaware,hu2024routerbenchbenchmarkmultillmrouting,chen2025panguembeddedefficientdualsystem} have emerged as the most representative paradigm that explicitly employs a routing mechanism to distribute tasks according to their difficulties, for instance, directing simple queries to LLMs or even small language models, while only keeping those complex problems for LRMs. Therefore, the challenge lies in developing an efficient router that enables optimal query distribution. RouteLLM~\cite{ong2025routellmlearningroutellms} leverages human preference data from Chatbot Arena to train a router with several algorithms. SoT~\cite{aytes2025sketchofthoughtefficientllmreasoning} defines three thinking paradigms and finetunes a DistilBERT as router. Similarly, \citet{li2025dynamicmindtrimodethinkinglarge} also introduce three think modes (i.e. fast, slow, normal) and train a classification model as router. ThinkSwitcher~\cite{liang2025thinkswitcherthinkhardthink} treats the routing problem as a regressing problem. Specifically, given a query, it samples the responses of the query under both slow-thinking and fast-thinking mode via prompt engineering, and then calculates the pass rates of the query for each mode. A regressor is then trained with the pass rates as supervision target and the embedding extracted from LRMs as features. Besides input difficulty, SwitchCoT~\cite{zhang2025longshortcotinvestigating} takes the token budget into consideration to select the best answer generation strategy. \citet{he2025selfrouteautomaticmodeswitching} believe that establishing routing mechanisms must account for model capability. If the general model or short-thinking mode can address the problem, that mode is activated; otherwise, LRM or long-thinking mode is triggered. Building on the findings from~\citet{ashok2025languagemodelspredictbehavior}, they train a classifier using the model's internal representation to probe capability boundary, thereby enabling capability-aware routing. \citet{liu2025tokensignaturepredictingchainofthought} observe that the monotonicity of token probability distributions may be correlated with the gains achieved through CoT reasoning. A logistic regression model is trained to adaptively apply CoT reasoning based on the spearman correlation, which measures the monotonic relationship between token probabilities and their sequence order. \citet{pan2025routereasonadaptiverouting} propose RTR, which is a unified routing framework to generate a routing table with both models and reasoning strategies given an input question. It represents candidate models and reasoning strategies to vectors. The input question is also encoded so that the expected performance and computational cost are predicted for each model-strategy combination.

\subsubsection{Other Pipeline}
CoThink~\cite{fan2025cothinktokenefficientreasoninginstruct} argues that preemptive difficulty assessment only based on query is unreliable, as many latent challenges emerge dynamically during the decoding stage. They propose a two-stage solution which uses an instruction model (e.g., Qwen2.5-Instruct) to generate concise outlines without really solving the problem, and then uses a reasoning model (e.g., DeepSeek R1) to complete it by following the high-density outlines. Similarly, Think-To-Think~\cite{zhao2025t2adaptivetesttimescaling} proposes to extract the reasoning strategies from similar questions, and select the best matched strategy for solving the given question. Inspired by speculative decoding, \citet{wang2025efficientreasoningllmsspeculative} propose Speculative Chain-of-Thought (SCoT), which leverages a light-weighted draft model (from the same family with the target model) to generate CoT drafts, the best CoT will be adopted for the target model to generate the final answer. The draft model is fine-tuned with very limited examples using LoRA to align the thinking behavior with the target model similar as CoT distillation. \citet{lin2025planbudgeteffectiveefficient} introduce a Bayesian Budget Allocation Model which reveals the relationship between token allocation and epistemic uncertainty. Based on this model, they propose a Plan-and-Budget framework which decomposes complex queries into sub-questions and allocates token budgets based on estimated complexity for each sub-query using adaptive scheduling. Several decay based scheduling strategies are applied. They argue that more tokens should be allocated to early sub-questions since epistemic uncertainty is typically highest at the beginning of reasoning process. \citet{lu2025prolongedreasoningneedcertaintybased} propose Certainty-based Adaptive Reasoning (CAR), which first generates a short answer and then evaluates the model's certainty of the answer by perplexity, long reasoning is only triggered when the model exhibits low confidence on this short answer.

Although these approaches can effectively reduce the reasoning length through well-designed pipelines, they also introduce additional inference overhead in modules other than the response generation module, which may result in increased end-to-end latency.

\subsection{Decoding Manipulation}
\label{sec:decoding_manipulation}

Decoding manipulation based approach refers to dynamically adjusting the decoding process to avoid redundant responses.

\subsubsection{Budget Forcing}
The simplest decoding manipulation approach is \textbf{budget forcing}~\cite{muennighoff2025s1simpletesttimescaling}, which forces a maximum and minimum number of reasoning tokens. Once the maximum reasoning token count reaches, ``\textit{</think>}'' is appended to early exit the thinking stage and generate the answer immediately. \citet{sun2025empiricalstudyllmreasoning} also propose to use keyword ``\textit{Time's Up}'' as the signal to stop reasoning when the reasoning token count reaches the budget. In contrast, if the reasoning tokens count does not reach the minimum count, the ``\textit{</think>}'' token will be bypassed and self-reflection tokens like ``\textit{Wait}'', ``\textit{Let me check}'' will be appended to foster reasoning continuously. As a result, LRMs are often guided to either overthink or underthink, and may even generate erroneous answers due to the abrupt truncation of their reasoning chains.



\subsubsection{Early Existing}
Instead of using a hard token budget, another line of decoding manipulation approaches is dynamic early exiting which exits the decoding process when the LRM is confident enough to generate the answer. Two fundamental questions that such approaches must address are: (1) \textit{when to trigger the early exit check}, and (2) \textit{what conditions or metrics should be met to perform early exit}. FlashThink~\cite{jiang2025flashthinkearlyexitmethod} simply checks the early exit at each decoding token, and directly prompt a LLM to verify whether the current reasoning pieces are enough to solve the problem. Although it can shorten reasoning chains, it also introduces heavy computational cost in frequently LLM verification. DEER~\cite{yang2025dynamicearlyexitreasoning} checks whether early exit can be done at each action transition point (e.g., ``\textit{Wait}'', ``\textit{Let me check}''). Upon encountering an action transition point, the model induces an answer from the existing reasoning pieces and evaluates the confidence using logits. Both answer induction and confidence evaluation are computed in parallel with the ongoing reasoning chain generation. So it will not introduce any additional latency. \citet{zhang2025reasoningmodelsknowtheyre,liu2025answerconvergencesignalearly} also show that probing the hidden states of LRMs can check the certainty of the model and verify intermediate answers with high accuracy. \citet{yong2025thinknotexploringthinking} check early exit at each reasoning step. The reasoning steps can be segmented by chunking techniques with syntactic cues or even LLMs. They also propose two metrics from an information theoretic perspective - \textit{InfoGain}, which captures the overall semantic alignment across the full reasoning output between the response and the golden output, and \textit{InfoBias}, which reflects the information gain (entropy reduction) of each reasoning step. During the decoding stage, the average entropy is calculated after each reasoning step, if it falls below the threshold, the LRM is prompted to directly generate the answer by appending ``\textit{</think>}'' tag to the think block if it is still within the thinking stage. \citet{liu2025answerconvergencesignalearly} check whether the model is sufficiently confident based on the consistency of the results from the $k$ consecutive reasoning steps. During decoding, the model is encouraged to generate the answer after each reasoning step. If the answers from the $k$ consecutive steps are consistent, it can be considered that the reasoning has converged. AlphaOne~\cite{zhang2025alphaonereasoningmodelsthinking} identifies the key limitations of existing approaches that fail to schedule slow thinking and fast thinking under overall token budget. They propose a unified perspective that firstly activates slow thinking as a Bernoulli stochastic process given a linear annealing scheduling function, and then terminates the reasoning by replacing any reasoning transition tokens like ``\textit{wait}'' with ``\textit{</think>}''. ConCISE~\cite{qiao2025conciseconfidenceguidedcompressionstepbystep} also applies a similar strategy to early exit reasoning when the model has generated the answer and is confident enough to bypass subsequent self-reflection steps. But the LRM should be fine-tuned to better align with such reasoning behavior, we will discuss this part in Section~\ref{sec:finetune_approaches}. \citet{fu2025efficientlyscalingllmreasoning} introduce a LRM serving system, which leverages the real-time computed certainty value to adjust the token budgets. 

\subsubsection{Token Logits Manipulation}
As we all know, most long CoTs contain many self reflection pieces, signaled by ``wait''-like tokens. Some of these tokens can induce self-correction or new insights, while some of them just affirm prior content. \citet{wang2025waitdontneedwait} and \citet{liu2025efficientreasoningsuppressionselfaffirmation} try to suppress such ``wait''-like tokens by adjusting the logits during decoding, thereby steering the model toward more efficient reasoning. Conversely, \citet{liu2025answerconvergencesignalearly} apply a linear transformation of the logit of \textit{``</think>''} token to encourage the model to terminate reasoning earlier. \citet{li2025steeringllmthinkingbudget} propose \textit{Budget Guidance}, which uses an auxiliary predictor to predict the remaining reasoning length probability distribution at each reasoning step. This distribution is then integrated with the original token probabilities to yield the final budget-conditional probability distribution.

\subsubsection{Activation Steering}
Recent studies have also revealed that the LRM reasoning process can be manually controlled via intervention such as activation steering~\cite{panickssery2024steeringllama2contrastive}. \citet{eisenstadt2025overclockingllmreasoningmonitoring} find that LRMs can monitor their relative position within the thinking process. In other words, the model knows how close it is to completing the reasoning. They train a linear regressor parametrized by $\theta$ to map the hidden representation in LRM to the relative positions of reasoning tokens. The parameter $\theta$, also called \textit{thinking progress vector}, is then used to control the decoding process by shifting the hidden representation. Similarly, \citet{sheng2025reasoningstrengthplanninglarge} believe that LRMs can preplan the reasoning strength before generation. Only a linear model can be trained to predict the number of reasoning tokens based on the hidden representation of \textit{``<think>''} token. Such pre-allocation ability could be obtained by \textit{pre-allocated direction vectors}. Based on this observation, they isolate such vectors using the difference-in-means method on questions of different difficulties. These vectors are then used to steer the hidden representation, thus controlling the reasoning strength. \citet{huang2025mitigatingoverthinkinglargereasoning} argue that such steering vector is not accurate enough and will introduce unintended interference noise. They further propose \textit{Manifold Steering} that projects the steering vector to a low-dimensional manifold. So the purified steering vector can be applied to more aggressively reduce the reasoning length with larger intervention strength.

\subsubsection{Parallel Scaling}
Unlike numerous existing works that conduct concise reasoning from a sequential parallel perspective,~\citet{hassid2025dontoverthinkitpreferring} explore whether a shorter response could achieve better performance through parallel scaling and propose an inference method called $short\text{-}m@k$. This method executes $k$ generations in parallel and halts the computation once the first $m$ thinking processes are done. Majority vote is used to induce the final answer among these $m$ chains. \citet{agarwal2025finishsearchefficienttesttime} adopt an even more radical strategy that directly returns the first finished trace as the final answer. \citet{wang2025thinkingshortrightthinking} design a serving system with a similar parallel scaling idea. They implement concurrent redundant sampling of responses beyond required, and terminate the sampling process upon reaching the required sampling count to avoid excessively long outputs. Simultaneously, they leverage a Process Reward Model (PRM) across sampling branches to prune those branches that are prone to generate inefficient responses, thereby further improving the sampling accuracy and releasing the computation resources for other promising branches.

\subsubsection{Other Decoding Manipulation}
\citet{dang2025internalbiasreasoningmodels} find that overthinking phenomenon is introduced by \textit{internal bias} of the model. The LRM tends to guess an answer before engaging in the reasoning process. So if the reasoning process is in conflict with the guess answer, it will get stuck in self-reflection. In order to validate this hypothesis, they simply mask the attention to the input question once the model first reaches an answer. This simple strategy can significantly reduce the length of the reasoning and even improve the accuracy in complex datasets. Similar to prompt-guided methods that add instructions in the prompt to make the model's output more concise, \citet{tang2025concisehintboostingefficientreasoning} propose ConciseHint which dynamically inserts text hint (e.g., ``make answer concise!'') during the decoding stage to make the model speak concisely, and adaptively adjusts the intervention strength and position according to input complexity.

Most decoding manipulation approaches design auxiliary modules to intervene in the decoding process, which allows for their seamless integration into existing inference systems, thereby rendering them non-invasive and widely applicable across multiple LRMs and tasks. 



\subsection{Model Merging}
\label{sec:model_merging}
Another line of training-free approaches to achieve adaptive thinking is model merging. Model merging combines a long-CoT model with a short-CoT model with simply weight averaging, aiming to create a ``medium-CoT'' model which can achieve balance between reasoning efficiency and accuracy. K1.5~\cite{kimiteam2025kimik15scalingreinforcement} shows that the model after model merging demonstrates superior token efficiency with only minimal performance degradation compared to the original LRM (k1.5-long). Given a high-accuracy model and a high-efficiency model, \citet{jin2025recutbalancingreasoninglength} use the DARE-Ties strategy~\cite{yu2024languagemodelssupermario} to perform linear interpolation on sparsely selected parameters. \citet{wu2025unlockingefficientlongtoshortllm} further conduct a comprehensive study on various model merging methods to achieve long-to-short reasoning. They find that model merging algorithms like task vector based methods~\cite{ilharco2023editingmodelstaskarithmetic} can deliver near 50\% token reduction alongside accuracy parity on 7B scale models. By employing a suitable calibration dataset, the activation-based model merging methods~\cite{nobari2025activationinformedmerginglargelanguage,liu2025sensmergingsensitivityguidedparameterbalancing} can achieve superior results in both performance and reasoning length reduction. They also reveal small-scale models (e.g., 1.5B) struggle to learn long CoT reasoning ability through model merging, and large scale models (e.g., 14B/32B etc.) pose significant challenges to balance the trade-off between performance and reasoning length. \citet{yao2025activationguidedconsensusmerginglarge} propose Activation-Guided Consensus Merging (ACM), which analyzes the mutual information between the activations of models under shared calibration corpus. The mutual information is then normalized as layer-specific weighting coefficients for model merging.

\subsection{Summary}
\label{sec:training_free_summary_anf_outlook}

The pursuit of training-free methods to mitigate excessive reasoning length in LRMs has emerged as a critical research frontier. This work categorizes these approaches into four paradigms: prompt-guided strategies, pipeline-based approaches, decoding manipulation techniques, and model merging. Prompt-guided methods exploit instruction-following capabilities via engineered prompts (e.g., direct prompt, token budgets, thinking mode, no-thinking directives). Although their simplicity enables rapid deployment, their efficacy hinges on the model's adherence to constraints, which is often inconsistent. Studies reveal unintended consequences, such as hidden inaccuracies and increased output instability, particularly when enforcing strict token limitations or suppressing reasoning steps. Pipeline-based methods modularize the reasoning workflow to reduce computational costs for LRMs through task offloading, while maintaining reasoning quality. Router-based approaches dynamically select optimal models/reasoning modes based on input complexity, model capabilities, or budget constraints. Other pipeline strategies include dynamic programming and iterative optimization as well as efficiency-enhancing techniques. These methods significantly shorten reasoning lengths but introduce auxiliary overheads (e.g., routing latency), leading to increased end-to-end delays, thus requiring a trade-off between efficiency and latency.

Decoding manipulation dynamically intervenes in the generation process through budget forcing, early exit checks, logit adjustments, or activation steering. Techniques like DEER~\cite{yang2025dynamicearlyexitreasoning} and FlashThink~\cite{jiang2025flashthinkearlyexitmethod} achieve shorter chains by monitoring confidence or semantic convergence, though frequent verification steps may offset computational savings. Parallel scaling strategies (e.g., $short\text{-}m@k$~\cite{hassid2025dontoverthinkitpreferring}) further enhance efficiency but require careful calibration to balance redundancy and accuracy. Model merging synthesizes long- and short-reasoning capabilities via parameter interpolation or activation-based fusion. Although effective for mid-scale models, this approach struggles with extreme scales (small or large models) and lacks fine-grained control over reasoning depth. Meanwhile, recent advances like Activation-Guided Consensus Merging (ACM)~\cite{yao2025activationguidedconsensusmerginglarge} highlight the potential of mutual information analysis in aligning heterogeneous models.  

\section{Training-based Methods}
\label{sec:training_based}

\subsection{Fine-tuning}
\label{sec:fine_tuning}
Enhancing the reasoning efficiency of LRMs through fine-tuning on variable-length CoT data has emerged as a promising direction. In this section, we review the existing literature, which can be broadly categorized into two interconnected stages: (1) \textbf{Data Construction}: this stage involves the synthesis of datasets featuring CoT reasoning paths of heterogeneous lengths. These datasets are typically generated using a variety of multifaceted techniques to ensure diversity in reasoning structure and complexity. (2) \textbf{Model Fine-tuning}: in this stage, the curated datasets are leveraged to train the models. Through methodologies like SFT or preference learning, models are trained to dynamically adjust the length of their reasoning, generating succinct chains for simple queries while providing more elaborate trajectories for complex ones. This adaptability ensures that the generated outputs remain both efficient and informationally sufficient.

\subsubsection{Data Construction}
\label{sec:data_construction}




Current studies have introduced various methodologies for constructing reasoning data. A notable example is ThoughtMani~\cite{liu2025thoughtmanipulationexternalthought}, which examines the natural traits of a LRM as it alternates between its ``thinking'' and ``non-thinking'' states. It leverages reasoning processes generated by a smaller non-reasoning model (CoT generator) and places them between thinking tokens. This efficient pipeline enables the LRM to directly assimilate essential information, thereby circumventing superfluous intermediate steps and substantially reducing computational overhead. Ada-R1~\cite{luo2025adar1hybridcotbileveladaptive} presents a bi-level preference data contruction strategy. For a given question, multiple solutions are sampled from both long and short reasoning models. Group-level preference allocation is determined by comparing these sampled responses. Subsequent intra-group comparisons are conducted to promote more concise reasoning, facilitating the development of instance-level preferences. Similarly, ReCUT~\cite{jin2025recutbalancingreasoninglength} proposes Stepwise Reasoning Trajectory Exploration mechanism, which iteratively expands a pool of diverse candidate trajectories. At each step, it prompts a LLM to generate multiple continuations of varying length, and then employs a reward function to prune the pool, selecting the optimal step for advancement. C3oT~\cite{kang2024c3otgeneratingshorterchainofthought} reduces the length of the original longer CoT while preserving essential information and interpretability. It trains LLM utilizing both the full-length and compressed CoT to understand the correlation between them. By leveraging the shorter CoT, C3oT captures the reasoning abilities derived from the longer CoT. A*-Thought~\cite{xu2025athoughtefficientreasoningbidirectional} represents the LRMs' reasoning process as a search tree, with nodes as reasoning spans. Employing the A* algorithm, it finds optimal paths by assessing ``information density'' and ``reasoning cost'' through a cost function, efficiently compressing the CoT.

Furthermore, TokenSkip~\cite{xia2025tokenskipcontrollablechainofthoughtcompression} operates on the principle that not all tokens hold equal importance in the reasoning process. It introduces a method for controlled CoT compression by strategically omitting less critical tokens, thereby enhancing computational efficiency without sacrificing logical integrity. To condition the model on a ``reasoning toggle'' instruction, \citet{bercovich2025llamanemotronefficientreasoningmodels} construct a dataset of paired examples. For each prompt, they generate two distinct outputs using specialized models: one response that includes a full reasoning chain and another that provides a direct, non-reasoning answer. Responses are labeled as ``detailed thinking on'' for reasoning and ``detailed thinking off'' for non-reasoning, prompting the model to adjust its reasoning based on the system prompt. While AutoL2S~\cite{luo2025autol2sautolongshortreasoning} combines comprehensive long CoT data for complex problems with concise short CoT data for simpler ones. By inserting the special <EASY> tags at strategic points within the short CoT pathways, it signals the model to bypass extensive reasoning where appropriate. This blended approach allows the model to learn, through training, how to identify ``simple problems'' and autonomously opt for shorter reasoning paths. To eliminate reliance on external controls, \citet{zhao2025letllmsbreakfree} propose Self-Braking Tuning (SBT), which allows models to self-regulate their reasoning process. SBT uses two data construction strategies: (1) SBT Exact (SBT-E), which precisely removes redundant segments based on predefined braking points, enabling earlier conclusions; (2) SBT Dynamic (SBT-D), which monitors each step and halts reasoning when overthinking is detected. While Nemotron-CrossThink~\cite{akter2025nemotroncrossthinkscalingselflearningmath} tackles the challenges of model generalization in reasoning by incorporating diverse datasets spanning STEM, humanities, and social sciences. By utilizing templates like multiple-choice and open-ended questions to control the complexity of the answer space, and by applying verifiable answer screening, it seeks to enhance the generalization capabilities of LLMs in various reasoning tasks. \citet{wu2025concisereasoningbiggains} observe the challenges of lengthy reasoning paths and inadequate adaptability in CoT distillation methods, leading to the proposal of Difficulty-Aware Prompting (DAP) to construct appropriate reasoning CoT data. DAP utilizes LLMs to assess problem difficulty and dynamically prunes reasoning paths accordingly, producing reasoning processes that are concise and complete. \citet{qiao2025conciseconfidenceguidedcompressionstepbystep} identify the issues of redundant reasoning as stemming from a confidence deficit and termination delay. In response, they introduce ConCISE, a methodology designed to generate superior reasoning CoT data by enhancing confidence through confidence injection, simplifying reasoning chains via Early Stopping, and circumventing unnecessary steps. To help the model understand its capability limits, \citet{he2025selfrouteautomaticmodeswitching} employ a dense complexity sampling strategy, which finely covers various levels of problem complexity. This approach ensures the dataset spans the full spectrum of difficulty, from simple tasks solvable with short CoT to complex challenges needing longer reasoning chains. Such balanced distribution allows the trained routing module to learn and identify features of different difficulty levels, enabling accurate mode selection for the model when addressing diverse problems. MoR~\cite{xiong2025mixturereasoningsteachlarge} leverages powerful LLMs like GPT-4o to generate a diverse pool of reasoning chain templates, which encapsulate various problem-solving strategies. These templates are then paired with problems from reasoning datasets. For each problem, a subset of templates is randomly selected, from which LLMs identify and choose the most suitable one. This optimal template is then combined with the problem to generate a full solution, complete with its reasoning process. The evaluation mechanism filters these solutions and only the correct responses are retained to construct the final training SFT dataset.

\begin{table*}[t]
\centering
\resizebox{0.6\textwidth}{!}{
\begin{tabular}{lc}
\hline
\textbf{Project}  & \textbf{Samples} \\ \hline
s1K-mix~\cite{yu2025longshortchainofthoughtmixturesupervised}  & 1,984             \\ \hline
OmniThought~\cite{cai2025reasoningomnithoughtlargecot} & 2000,000         \\ \hline
Dolphin-r1~\cite{dolphin-r1} & 800,000         \\ \hline
OpenThoughts-114k~\cite{guha2025openthoughtsdatarecipesreasoning} & 114, 000    \\ \hline
OpenMathReasoning~\cite{moshkov2025aimo2}  & 306,000      \\ \hline
ThinkPatterns-21k~\cite{wen2025thinkpatterns21ksystematicstudyimpact} & 21,000           \\ \hline
Llama-Nemotron~\cite{bercovich2025llamanemotronefficientreasoningmodels} & 33,011,757              \\ \hline
\end{tabular}}
\caption{Summary of open-sourced variable-length CoT data.}
\label{table:var_len_cot_data}
\end{table*}

\subsubsection{Variable Length Dataset}
\label{sec:variable_length_dataset}

The construction of specialized datasets serves as a key strategy for advancing reasoning models. This approach involves creating tailored data to instill specific reasoning capabilities. \citet{cai2025reasoningomnithoughtlargecot} identify a lack in large-scale coherent CoT problems drawn from multiple teacher models, and note that existing datasets neglect attributes essential for LRMs development. Then they propose the OmniThought dataset, which comprises 2 million CoT reasoning processes, and generated and validated by two powerful LRMs serving as teacher models. They also propose two unique metrics: Reasoning Veracity~(RV) for CoT thoroughness and Cognitive Difficulty~(CD) for evaluating model comprehension challenges. \citet{zhao202514millionopensourcedistilled} present a large-scale dataset called AM-DeepSeek-R1-Distilled, comprising 1.4 million samples focused on the cognitive processes central to general reasoning tasks. Curated from diverse open-source datasets, it undergoes semantic deduplication and thorough cleansing to avert testset contamination. Responses are distilled from reasoning models, and subjected to rigorous verification to guarantee accuracy and reliability. To further enhance the capacity of LRMs for automatic hybrid reasoning, \citet{zhang2025othinkr1intrinsicfastslowthinking} explicitly construct a SFT dataset. For tasks where fast-thinking LLMs yield correct answers and the reasoning trajectories of LRMs are judged redundant, they remove the redundant reasoning trajectories, retaining only the immediate responses. In addition, guided by the \textit{Thinking then Responding} paradigm, \citet{wen2025thinkpatterns21ksystematicstudyimpact} examine the effects of different cognitive process types on model performance. They introduced five internal thinking modes to each data pair: one unstructured mode (i.e. monologue) and four structured modes (i.e. decomposition, self-questioning, self-debate, and self-critique), while keeping the instructions and responses unchanged. This collection constitutes the ThinkPatterns-21k dataset, consisting of 21,000 instruction-response pairs.

In a complementary effort, the focus on data quality and complexity has led to the development of challenging, domain-specific benchmarks. DeepMath-103K~\cite{he2025deepmath103klargescalechallengingdecontaminated} is an extensive dataset curated to facilitate advanced reasoning in mathematics and interdisciplinary domains. It consists of high-difficulty problems (levels 5-9) that have been meticulously decontaminated to ensure integrity. Each problem is accompanied by verifiable solutions and includes three distinct Deepseek R1 solution variants to support diverse training methodologies. s1~\cite{muennighoff2025s1simpletesttimescaling} curated the s1K dataset, consisting of 1,000 questions and reasoning paths, through a detailed three-stage filtering process focused on difficulty, diversity, and quality. The initial pool comprised from 16 sources, ensuring quality and challenge across various fields. After data cleaning and model-based filtering, the dataset was refined to a manageable size. The questions were sorted into 50 domains using the Mathematics Subject Classification system, prioritizing those with longer reasoning traces to highlight complexity. \citet{tian2025deepdistillenhancingllmreasoning} develop a high-quality dataset through the application of multi-model distillation and systematic difficulty grading, aimed at addressing a wide array of reasoning challenges. The dataset comprises 3.34 million unique queries and 40 million distilled responses spanning fields such as mathematics, coding, and science. The difficulty levels are rigorously quantified using pass rates along with the coefficient of variation, which evaluates the stability of responses to repeated queries.

Table~\ref{table:var_len_cot_data} summarizes open-source variable-length CoT data. These datasets either contain data from non-reasoning domains or contain CoT data with variable length.

\subsubsection{Fine-tuning Approaches}
\label{sec:finetune_approaches}





After constructing and collecting variable-length reasoning data, the focus of current research is on the development of effective fine-tuning methodologies. This involves optimizing models to handle diverse reasoning pathways and lengths efficiently, ensuring that they can adapt to varying complexities and improve their overall reasoning performance. Researchers are especially interested in finetune techniques that enhance model generalization, maintain high accuracy, and take advantage of the richness of the variable-length dataset to improve adaptive reasoning. 
 
\textbf{Long CoT Compression Fine-tuning.} Strategies for fine-tuning models on long CoT compression primarily fall into three categories: data-centric curriculum design, dynamic inference path modification, and structured multi-stage reasoning.

A foundational approach is data-centric, where models are explicitly taught to generate shorter reasoning chains by curating or creating specialized data. For instance, CoT-Valve~\cite{ma2025cotvalvelengthcompressiblechainofthoughttuning} enables models to generate reasoning chains of varied lengths by constructing datasets with chains ranging from long to short for identical questions, while investigating precise length-compressible CoT tuning and progressive chain length compression methods. Similarly, \citet{yu2025longshortchainofthoughtmixturesupervised} propose LS-Mixture SFT, which integrates long CoT inference data with short CoT data created through structure-preserving rewriting, directly training the model to handle both elaborate and concise reasoning.

Another line of work focuses on dynamic inference mechanisms. System-1.5 Reasoning~\cite{wang2025system15reasoningtraversallanguage} reveals dynamic shortcut mechanisms within latent spaces through model depth shortcuts (DS) and step shortcuts (SS), achieving this by distilling natural language CoT into enduring cognitive sequences, further refining System-2 latent reasoning into adaptive pathways. This principle of dynamic adaptation is also central to Adaptive GoGI-Skip~\cite{zhuang2025acceleratingchainofthoughtreasoninggoalgradient}, which employs goal-oriented gradient importance and dynamic skipping mechanisms for CoT compression, thereby achieving high compression level while preserving reasoning fidelity. 



Finally, some methods achieve compression through structured, multi-stage reasoning frameworks that mimic human cognitive flexibility. A prominent example is Thinker~\cite{chung2025thinkerlearningthinkfast}, which decomposes the QA task into four stages (that is, Fast Thinking, Verification, Slow Thinking, and Summarization) with defined objectives and reward mechanisms, allowing for process termination upon successful verification, or alternatively, transitioning to Slow Thinking. Each stage involves specific token budget constraints to improve the efficiency and accuracy of task completion.

\textbf{Short CoT Selection Fine-tuning.} Another research direction focuses on fine-tuning models to selectively generate or identify concise reasoning paths from the outset. These methods often leverage self-improvement loops, verification mechanisms, or RL-style fine-tuning.

A notable strategy is training models on self-generated, high-quality concise reasoning paths. \citet{munkhbat2025selftrainingelicitsconcisereasoning} employ best-of-N sampling alongside few-shot conditioning to produce shorter reasoning processes, which are then used as a fine-tuning dataset to enhance the model's capacity for succinct reasoning. Building on the same principle of self-improvement, \citet{yang2025thinkingoptimalscalingtesttimecompute} introduce Thinking-Optimal Scaling which begins with seed data diverse in response length and trains the model to apply varying levels of reasoning effort. The model subsequently self-improves by identifying its shortest correct responses across various problems, adapting effectively to different reasoning requirements.

A different way achieves efficiency not by shortening every path, but by enabling the model to determine when to terminate. VeriThinker~\cite{chen2025verithinkerlearningverifymakes} incorporates auxiliary verification tasks, empowering models to accurately self-assess the correctness of current CoT solutions. This enables autonomous decision-making regarding the necessity of further self-reflection steps, thus reducing instances of overthinking.

Furthermore, RL-style fine-tuning optimizes the trade-off between conciseness and accuracy. O1-Pruner~\cite{luo2025o1prunerlengthharmonizingfinetuningo1like} utilizes RL-style fine-tuning to promote concise reasoning chains. By first establishing a performance baseline via pre-sampling, the method ensures that the pursuit of conciseness does not compromise adherence to strict accuracy constraints.

\textbf{Implicit CoT Fine-tuning.} Implicit CoT fine-tuning seeks to internalize the reasoning process, enabling models to ``think'' efficiently without generating verbose textual steps. ~\citet{pfau2024letsthinkdotdot} observe that even inserting semantically meaningless tokens (e.g., ``...'') can enhance reasoning performance, suggesting that the mere allocation of additional computational budget is beneficial. This insight motivated a broader inquiry into how to make the allocated ``thinking time'' more structured and meaningful. A central strategy is to replace explicit CoT with compact latent representations. Although existing survey~\cite{zhu2025surveylatentreasoning} has summarized methods for transforming CoT into latent reasoning, our survey still provides a systematic review and synthesis of approaches related to implicit CoT fine-tuning. Early approaches focused on compressing reasoning into single, high-level tokens, as shown in LightThinker~\cite{zhang2025lightthinkerthinkingstepbystepcompression} and Heima~\cite{shen2025efficientreasoninghiddenthinking}. This concept was extended to variable-length sequences of continuous ``contemplation tokens'' by CCoT~\cite{cheng2024compressedchainthoughtefficient}. More dynamic methods have also been proposed. For example, CoLaR~\cite{tan2025thinksilentlythinkfast} achieves adaptive dynamic compression via an auxiliary task ``next compressed embedding prediction'' and a dedicated latent head in the latent space. Others leverage hybrid representations: \citet{su2025tokenassortedmixinglatent} abstract initial reasoning steps into latent discrete tokens by VQ-VAE and blend them with standard text tokens during training. In parallel, Coconut~\cite{hao2024traininglargelanguagemodels} leverages the final hidden state of the LLM as a ``continuous thought'' representation of the reasoning state, directly feeding it back to the model as subsequent input embeddings in the continuous latent space.

Another powerful paradigm is knowledge distillation, where a student model learns to replicate the outcome of a teacher's explicit CoT process. A straightforward implementation is the ``progressively decrease CoT'' strategy~\cite{deng2024explicitcotimplicitcot}, which gradually weans the model off its dependency on explicit steps. The final model can derive answers without outputting any CoT steps, while implicitly embedding the reasoning process within. This is formalized by \citet{yu2024distilling21}, who distill the capabilities of a deliberative ``System 2'' (CoT) into a reactive ``System 1'' (the base LLM) to directly acquire the capability of ``skipping intermediate steps to output correct answers'' via self-supervised learning. A more sophisticated variant is self-distillation, exemplified by CODI~\cite{shen2025codicompressingchainofthoughtcontinuous}, where the model acts as its own teacher, learning to align the hidden states of its explicit and implicit reasoning paths on critical answer-generating tokens.

Finally, some research pursues implicit CoT fine-tuning by introducing architectural innovations. \citet{liu2024expeditingelevatinglargelanguage} use a auxiliary model to compress complete reasoning chains into semantically aligned compact tokens to fine-tune the primary model. While SoftCoT~\cite{xu2025softcotsoftchainofthoughtefficient} uses a lightweight model to generate instance-specific ``soft reasoning tokens'', avoiding modification to the base LLM. A more fundamental architectural change is iterative processing. \citet{saunshi2025reasoninglatentthoughtspower} introduce ``Looped Transformers'', a method that dynamically adjusts reasoning depth by varying the loop iterations. It enables fine-tuning to tailor reasoning complexity to different tasks through loop adjustments. Building directly on this, \citet{yu2025enhancingautoregressivechainofthoughtloopaligned} further propose RELAY, which aligns CoT reasoning steps with loop iterations and utilizes the structure to generate accurate reasoning chains for complex problems exceeding training length, long-form reasoning data for further fine-tuning.

\textbf{DPO Variant Fine-tuning.} Recent works adapt preference optimization algorithms, such as DPO variants, to fine-tune reasoning processes by navigating the trade-off between conciseness and accuracy through novel preference schemes.

A foundational strategy is to unambiguously prioritize length. \citet{su2025underthinkingoverthinkingempiricalstudy} adapt the SimPO algorithm to prioritize response length as the primary supervision signal. By consistently preferring the shorter of two sampled responses—irrespective of its accuracy—this method enables training on unlabeled data and significantly reduces the computational cost of data generation. Some approaches seek a more nuanced balance to avoid sacrificing accuracy. For instance, \citet{chen2025think23overthinkingo1like} build upon SimPO with a similar length-preference foundation but integrate a First Correct Solution (FCS) strategy. This involves discarding redundant reasoning that appears after the first correct answer. To mitigate the risk of losing reasoning depth, the framework also retains a second correct solution, thereby preserving the model's ``reflection'' capabilities. Other methods achieve this balance through more explicit multi-objective frameworks. ReCUT~\cite{jin2025recutbalancingreasoninglength} trains two distinct preference models: an accuracy model on ``shortest-correct'' vs. ``longest-incorrect'' trajectories and a length-optimization model on ``shortest-correct'' vs. ``longest-correct'' trajectories. By interpolating the parameters of these two models, ReCUT produces a fused model that judiciously balances both accuracy and conciseness. \citet{an2025dontthinklongerthink} introduce Dynamic Thinking Pattern Optimization (DTO), which segments model-derived reasoning trajectories into distinct patterns and evaluates each segment's contribution to overall efficiency. This segmentation allows for the pruning of negative segments and the reinforcement of positive ones through a preference optimization mechanism that utilizes pairwise datasets. In addition, a distinct strategy frames the problem around adherence to external constraints, such as token budgets. TALE-PT~\cite{han2025tokenbudgetawarellmreasoning} first identifies an optimal token budget for a given query. It then employs a two-stage process: first, an SFT phase to teach the model to align with the budget, followed by a DPO-based internalization phase that reinforces refined reasoning paths that satisfy the budget criteria. This strategy empowers the model to generate concise and effective reasoning outputs without necessitating extensive labeled datasets.

\textbf{Other Fine-tuning.} Researchers have developed a diverse range of fine-tuning strategies to enhance model's adaptability and reasoning capability. A dominant theme is the explicit integration of dual cognitive systems, inspired by ``fast'' (System 1) and ``slow'' (System 2) thinking.

This integration is often achieved through sophisticated data-centric approaches. Dualformer~\cite{su2025dualformercontrollablefastslow} constructs a mixed training dataset of both ``full reasoning chains'' and ``result-oriented'' strategies, using structured dropout techniques to mimic cognitive shortcuts. Others use explicit instructions to trigger different reasoning modes; \citet{wang2025adaptivedeepreasoningtriggering} fine-tune on both short and long CoT datasets, applying both original and augmented data with instructions like ``Please answer with Long/Short CoT'' to tailor reasoning modes effectively. A more dynamic approach is proposed by \citet{li2025tldrlongreweightingefficient}, who use a dynamic ratio of System 1 and System 2 data during training to optimize the model's adaptive balance.

Other methods pursue fine-tuning optimization by introducing novelty to the loss function and training pipeline for adaptive reasoning. \citet{zhang2025othinkr1intrinsicfastslowthinking} design a loss function with dual-reference KL-divergence constraints, enabling the model to enhance dynamic switching between cognitive modes and learn output distributions from both ``fast-'' and ``slow-'' thinking reference simultaneously. Qwen3~\cite{yang2025qwen3technicalreport} presents a multi-stage pipeline that begins with SFT on long CoT, followed by RL. It trains and culminates in fusing this ``non-thinking'' functions with the established ``thinking'' model. \citet{tencenthunyuanteam2025hunyuanturbosadvancinglargelanguage} train Hunyuan-Base with reasoning data to obtain a short CoT model. After conducting consistency checks, they use Hunyuan-T1 to generate extended reasoning for incorrect samples and convert them into short CoT style, then concatenate all failed attempts with correct responses to form complete training data, enabling the model to both learn the efficiency of short CoT and absorb the reasoning capabilities of long CoT.

A stream of work reimagines the structure of the reasoning process. ~\citet{wang2024guidinglanguagemodelreasoning} introduce a hierarchical generation scheme in which the LLM initially produces a ``planning token'' at the start of each reasoning step. These tokens intuitively function as high-level plans for the current step, and their embeddings are integrated into the model parameters.  Similarly, TH2T~\cite{liu2025thinkthinkmitigatingoverthinking} introduces a ``self-hypnosis'' mechanism, using specialized tokens (e.g., ``difficulty-hypnosis'' token and ``redundancy hypnosis'' token) to modulate the reasoning process by dynamically managing perceived task difficulty and step redundancy. MinD~\cite{zeng2025betterperfectunlockingefficient}, in contrast, reframes reasoning as an interactive, structured multi-turn sequence, allowing users to reflect, verify or explore at each step, transforming a monolithic process into a manageable, iterative one.

Some methods explore unconventional fine-tuning paradigms. \citet{liu2025qfftquestionfreefinetuningadaptive} propose Question-Free Fine-Tuning (QFFT), which surprisingly removes input questions from the training data, fine-tuning the model solely on Long CoT responses. This encourages the model to learn the reasoning patterns themselves, breaking the rigid question-response mapping and preserving its inherent fast-thinking capabilities. These diverse approaches collectively underscore the sophisticated methodologies employed in the fine-tuning of LLMs for improved reasoning and adaptability.

\subsection{Reinforcement Learning}
\label{sec:rl}



The effectiveness of Reinforcement Learning in improving reasoning capability has been demonstrated by multiple recent studies. However, a growth of studies has shifted focus toward the issue of token redundancy, aiming to reduce token costs while preserving strong reasoning capabilities, thereby enabling true adaptive reasoning. Existing research has comprehensively explored adaptive reasoning from various different aspects. In this section, we classify these methods mainly into five categories: \textbf{RL with Length Penalty}, \textbf{RL with GRPO-Variant}, \textbf{RL with Difficulty-awareness}, \textbf{RL with Thinking Mode}, and \textbf{Other RL}. As this survey aims to comprehensively classify by different aspects, some overlap exists among these categories.


\begin{table*}[t]
\centering
\resizebox{1\textwidth}{!}{
\begin{tabular}{lllll}
\hline
Method                                                                  & RL Algorithm                  & On/Off Policy                 & \multicolumn{1}{c}{Reward Function} \\ \hline
~\citet{arora2025traininglanguagemodelsreason}          & \multicolumn{1}{c}{PPO}       & \multicolumn{1}{c}{On Policy} & \multicolumn{1}{c}{$\begin{cases} 
1-\alpha\cdot\sigma\left(\frac{L(y)-\mathrm{MEAN}(L)}{\mathrm{STD}(L)}\right) & \mathrm{if~}S(y)=1 \\ 
0 & \mathrm{if~}S(y)=0 
\end{cases}$}                       \\
\hdashline
~\citet{ling2025fasteasydeephard}                      & \multicolumn{1}{c}{REINFORCE} & \multicolumn{1}{c}{On Policy} & \multicolumn{1}{c}{$\begin{cases} 1+\frac{\alpha}{L(y)^\gamma} & \mathrm{if~}S(y)=1 \\ 0 & \mathrm{if~}S(y)=0 \end{cases}$}                  
\\
\hdashline
~\citet{yeo2025demystifyinglongchainofthoughtreasoning} & \multicolumn{1}{c}{PPO}       & \multicolumn{1}{c}{On Policy} & \multicolumn{1}{c}{$\quad \begin{cases} \mathrm{CosFn}(L(y),L_{\mathrm{max}},r_{0}^{c},r_{L}^{c}) & \mathrm{if~}S(y)=1 \\ \mathrm{CosFn}(L(y),L_{\mathrm{max}},r_{0}^{w},r_{L}^{w}) & \mathrm{if~}S(y)=0 \\ r_{e} & \mathrm{if~}L(y)=L_{\mathrm{max}} \end{cases}$} \\
\hdashline
~LCPO\cite{aggarwal2025l1controllinglongreasoning} & \multicolumn{1}{c}{GRPO}       & \multicolumn{1}{c}{On Policy} & \multicolumn{1}{c}{$S(y)-\alpha\cdot|L_{budget}-L(y)|$} \\
\hdashline
~\citet{su2025thinkingfastrightbalancing} & \multicolumn{1}{c}{GRPO}       & \multicolumn{1}{c}{On Policy} & \multicolumn{1}{c}{$S(y)-\lambda_t\cdot L(y)$} \\
\hdashline
~DAST~\cite{shen2025dastdifficultyadaptiveslowthinkinglarge} & \multicolumn{1}{c}{SimPO}       & \multicolumn{1}{c}{Off Policy} & \multicolumn{1}{c}{$\begin{cases}
\max(-0.5\cdot \frac{L(y) - L_{budget}}{L_{budget}} + 0.5, 0.1) & \mathrm{if~}S(y)=1 \\
\min(0.9\cdot \frac{L(y) - L_{budget}}{L_{budget}} - 0.1, -0.1) & \mathrm{if~}S(y)=0
\end{cases}$} \\
\hdashline
~DAPO~\cite{yu2025dapoopensourcellmreinforcement} & \multicolumn{1}{c}{GRPO}       & \multicolumn{1}{c}{On Policy} & \multicolumn{1}{c}{$\begin{cases} 
S(y) & \mathrm{if~} L(y)\leq L_{\mathrm{max}}-L_{\mathrm{cache}} \\
S(y) + \frac{L_{\mathrm{max}}-L_{\mathrm{cache}}-L(y)}{L_{\mathrm{cache}}} & \mathrm{if~} L_{\mathrm{max}}-L_{\mathrm{cache}}<L(y)\leq L_{\mathrm{max}} \\
S(y) -1 & \mathrm{if~} L_{\mathrm{max}}<L(y) 
\end{cases}$} \\
\hdashline
~\citet{song2025walkrunconcisellm}  & \multicolumn{1}{c}{GRPO}       & \multicolumn{1}{c}{On Policy} & \multicolumn{1}{c}{$\begin{cases} 1+\lambda\left(1-\frac{L(y_i)}{L_{\max}}\right) & \mathrm{~if~}\sum_{j=1}^GS(y_j)=G \\ S(y_i) & \mathrm{~if~}\sum_{j=1}^GS(y_j)\neq G \end{cases}$} \\
\hdashline
~Short-RL~\cite{yuan2025efficientrltrainingreasoning} & \multicolumn{1}{c}{PPO}       & \multicolumn{1}{c}{On Policy} & \multicolumn{1}{c}{$\begin{cases} 0.5-\frac{L(y)-L_{\min}}{L_{\max}-L_{\min}} & \mathrm{if~}S(y)>0\mathrm{~and~acc\geq acc_{\max}-\tau_{acc}~and~}L(y)>L_{\min}+\tau_{\ell} \\ 0.5 & \mathrm{if~}S(y)>0\mathrm{~and~acc\geq acc_{\max}-\tau_{acc}~and~}L(y)\leq L_{\min}+\tau_{\ell} \\ 0 & \mathrm{otherwise} \end{cases}$} \\
\hdashline
~ASRR~\cite{zhang2025continuethinkingadaptivethinking} & \multicolumn{1}{c}{PPO}       & \multicolumn{1}{c}{On Policy} & \multicolumn{1}{c}{$\begin{cases} S(y) & \mathrm{if~Acc}_G<\tau \\ S(y)-\frac{\beta\cdot(\mathrm{Acc}_G-\tau+\epsilon)}{1-\tau+\epsilon}\cdot\min\left(1,\max\left(0,\frac{L(y)-L_{\min}}{L_{\mathrm{window}}}\right)\right) & \mathrm{if~Acc}_G\geq\tau \end{cases}$} \\
\hdashline
~HAPO~\cite{huang2025hapotraininglanguagemodels} & \multicolumn{1}{c}{GRPO}       & \multicolumn{1}{c}{On Policy} & \multicolumn{1}{c}{$\begin{cases} S(y) & \mathrm{if~}L_\mathrm{hist}\mathrm{~is~Null} \\ 1+w\cdot\max\left(\cos\left(\min\left(\frac{\pi}{2}\frac{L(y)}{L_\mathrm{hist}},\pi\right)\right),c\right) & \mathrm{if~}L_\mathrm{hist}\text{ exists and }S(y)=1 \\ w\cdot\min\left(\cos\left(\min\left(\frac{\pi}{2}\frac{L(y)}{L_\mathrm{hist}},\pi\right)\right),0\right) & \mathrm{if~}L_\mathrm{hist}\text{ exists and }S(y)=0  \end{cases}$} \\
\hline
\end{tabular}}
\caption{Comparison of selected reward function for adaptive reasoning in LRMs. $S(y)\mathrm{~\in~0,1}$ evaluates the correctness of the generated answer, and $L(y)$ represents the output length.}
\label{table:reward_function_rl}
\end{table*}

\subsubsection{RL with Length Penalty}
\label{sec:rl_with_length}



A primary strategy to strengthen reasoning efficiency involves incorporating a length penalty into the RL framework. These methods aim to curtail verbose or redundant steps, but vary significantly in how to define and apply the length penalty to balance conciseness with accuracy. Table~\ref{table:reward_function_rl} summarizes part of the reward functions for adaptive reasoning in LRMs.

A foundational direction focuses on shaping the reward function to be length-aware. The simplest form, shown in \citet{arora2025traininglanguagemodelsreason}, directly ranks the reward scores of the sampled per-prompt solutions according to their correctness and normalized length, and always keeps shorter correct answers higher. However, recognizing that complex problems require longer reasoning, many works have developed more sophisticated, non-linear penalties. For instance, \citet{ling2025fasteasydeephard} adopt a powered length penalty (PLP) that is more lenient on longer responses to allow for necessary reasoning on difficult questions, while \citet{yeo2025demystifyinglongchainofthoughtreasoning} introduce a sparse cosine length-scaling reward to stabilize CoT growth. Others modulate the reward dynamically; Autothink~\cite{tu2025learningthinkshapingadaptive} employs a length-aware reward modulation where rewards decrease with the length of correct responses and increase for incorrect ones, promoting concise success and comprehensive failure analysis. Similarly, DAPO~\cite{yu2025dapoopensourcellmreinforcement} applies a soft overlong punishment to mitigate reward noise resulting from improper shaping of the truncated samples, escalating punishment only after a set length is exceeded; and \citet{wang2025adaptivethinkingmodepolicy} design a smooth penalty function suited for multi-turn dialogues that moderates the difference between actual and desired response lengths.

Beyond the shape of the reward function, another key strategy involves conditionally applying the penalty to protect reasoning quality. The core idea is to avoid penalizing length when it might harm accuracy. Short-RL~\cite{yuan2025efficientrltrainingreasoning} exemplifies this by applying penalties only to correct answers or deactivating them entirely if batch accuracy drops. ASRR~\cite{zhang2025continuethinkingadaptivethinking} follows a similar principle, imposing penalties only after a target accuracy is met, thereby balancing efficiency and correctness. Likewise, Light-R1~\cite{wen2025lightr1curriculumsftdpo} limits the shortening reward for correct answers to prevent a "length collapse" early in the training. \citet{su2025thinkingfastrightbalancing} also integrate a dynamic adjustment mechanism into the normal length penalty to achieve dynamic penalty, where the penalty coefficient is adjusted in real-time based on the model's accuracy rate.

A distinct line of work frames length optimization around adherence to external constraints, such as token budgets or explicit targets. LCPO~\cite{aggarwal2025l1controllinglongreasoning} designs an explicit mechanism to obey the length restrictions specified in the prompt (i.e., ``Think for $n_{gold,i}$ tokens''), and proposes a dual-purpose reward function with a length penalty to encourage the response brief but meet the target length. Several methods are budget-aware: DAST~\cite{shen2025dastdifficultyadaptiveslowthinkinglarge} applies budget-aware reward shaping to penalize those responses exceeding the token budget of simple questions and courage the harder ones approaching the metric; while \citet{song2025walkrunconcisellm} reward models for leaving more of their budget unused. Specifically, When rollout results for a problem are all correct, they use the maximum remaining response length as the length reward, where more remaining context length results in a greater reward. \citet{qi2025optimizinganytimereasoningbudget} develop an AnytimeReasoner framework, optimizing reasoning efficiency under variable token budgets by sampling from a prior distribution and enforcing dense rewards through truncated yet verifiable reasoning. The method decouples reasoning from summarization policy learning and introduces Budget Relative Policy Optimization to stabilize advantage estimation.

Besides, some approaches embed length penalties within advanced, multi-component systems. THINKPRUNE~\cite{hou2025thinkprunepruninglongchainofthought} compresses reasoning through RL training with strict token limits, truncating and penalizing outputs that exceed these limits. An iterative pruning schedule progressively tightens length constraints, selecting the most concise model within acceptable performance tolerance for successive rounds.  HAPO~\cite{huang2025hapotraininglanguagemodels} presents a history-aware length reward function designed to encourage finding correct solutions that are more concise than prior ones. This reward function aims to minimize excessive penalties on shorter incorrect answers, thereby promoting exploration of more efficient solutions. REA-RL~\cite{deng2025rearlreflectionawareonlinereinforcement} combines two components: a lightweight reflection model for multi-path reasoning that prunes redundancy by pinpointing the first correct answer, thus refining the reasoning sequence; and a reflection-aware reward mechanism penalizing inadequate reflection via keyword density analysis, alongside a modified length reward to curtail unnecessary expansion.

\subsubsection{RL with GRPO-Variant}
\label{sec:rl_with_grpo}

GRPO is a powerful RL approach for training models to adapt different reasoning modes, but its standard form can lead to issues like ``Format Collapse''—an overreliance on the single most accurate reasoning format. Consequently, numerous studies have developed GRPO variants that introduce more sophisticated control mechanisms, primarily by addressing format diversity, integrating external guidance, or redesigning the loss function.

To tackle the "Format Collapse" problem, Ada-GRPO~\cite{wu2025armadaptivereasoningmodel} addresses this problem by boosting the reward for less frequently sampled reasoning formats, ensuring they receive sufficient learning opportunities. A more structural solution is DeGRPO (Decoupled GRPO)~\cite{fang2025thinklessllmlearnsthink}, which decomposes the learning objective into two parts: a mode selection loss to optimize the choice of reasoning mode and a response loss to focus on answer accuracy.

Another major method is to enhance GRPO with external guidance. Several methods are budget-aware: SelfBudgeter~\cite{li2025selfbudgeteradaptivetokenallocation} uses a budget-guided GRPO to align response length with problem difficulty, The designed reward function centers on three key aspects: answer correctness, minimal token usage, and alignment between response length and allocated token budget; while Elastic Reasoning~\cite{xu2025scalablechainthoughtselastic} explicitly divides reasoning into two phases—thinking and solution—with separately allocated budgets. To enhance robustness to truncated reasoning, Elastic Reasoning incorporates GRPO with a budget-constrained rollout strategy, enabling the model to adaptively reason under shortened thinking processes. Others are difficulty-aware: DIET~\cite{chen2025overthinkersdietcuttingtoken} integrates real-time difficulty estimation within the RL framework, allowing dynamic adjustment of token compression policies to optimize performance and efficiency; AdaCtrl~\cite{huang2025adactrladaptivecontrollablereasoning} employs a multi-objective reward that combines accuracy, difficulty calibration, and difficulty-aware length control.

Some variants introduce fundamental innovations to the reward and optimization function itself. Some propose composite reward systems; ACPO~\cite{cheng2025incentivizingdualprocessthinking} employs a composite reward system comprising accuracy, online Token Length Budget (TLB), and system mode rewards. This design helps the model optimize reasoning by balancing accuracy with efficiency and cognitive mode appropriateness. HGPO~\cite{jiang2025thinkneedlargehybridreasoning} employ a rule-based assignment scheme to normalize scores, jointly capturing both the relative quality across reasoning modes and the answer quality within each individual mode through inter-group and intra-group rewards. ConciseR~\cite{song2025walkrunconcisellm} has optimized GRPO, producing two improved versions: GRPO++ and L-GRPO. GRPO++ integrates DAPO's core technologies with entropy rewards, while L-GRPO revises the length reward function, keeps the KL divergence penalty, and uses entropy rewards to sustain exploration. LC-R1~\cite{cheng2025optimizinglengthcompressionlarge} uses a dual reward system to both encourage overall compression and prune specific invalid reasoning segments. CoLaR leverages the non-determinism of its Latent Head to generate diverse reasoning paths. These paths are subsequently refined by a GRPO-variant algorithm (without KL regularization), thereby reinforcing reasoning chains that are both correct and more compact. Besides, \citet{ding2025thinkingtokenshelptrap} introduce Dual Policy Preference Optimization (DuP-PO), which extends the GRPO framework through innovations such as Dual-Policy Sampling, Token-Level Advantage Scaling, and Policy Shaping, to suppress overthinking triggered by thinking tokens (e.g., wait, however) in large reasoning models and achieve a balanced improvement in reasoning efficiency and performance.

\subsubsection{RL with Difficulty-awareness}
\label{sec:rl_with_difficulty}


Various studies have investigated adaptive response length adjustments, leveraging difficulty-awareness in trained models—ensuring succinct answers for simple questions and encouraging more detailed reasoning for complex ones to boost accuracy. The primary distinction among these methods lies in how they estimate or infer this difficulty.

The most direct approach involves training the model to explicitly estimate problem difficulty or a corresponding token budget. For instance, DAST~\cite{shen2025dastdifficultyadaptiveslowthinkinglarge} introduces a TLB metric to align target response length length with problem complexity, while SelfBudgeter~\cite{li2025selfbudgeteradaptivetokenallocation} mitigates overthinking by predicting the minimal token budget needed according to problem complexity, thereby reducing user waiting time. More dynamic systems, such as DIET~\cite{chen2025overthinkersdietcuttingtoken} and AdaCtrl~\cite{huang2025adactrladaptivecontrollablereasoning}, integrate this difficulty estimation directly into the RL loop. Taking model-based difficulty estimation with adaptive weighting with dynamic length targets, DIET guarantees concise outputs for straightforward problems and thorough reasoning for intricate ones; AdaCtrl employs multiple rollouts from RL training to refine its difficulty judgment, enabling more efficient reasoning resource allocation. For complex problems, the length initially increases and then stabilizes at a longer level. For simple problems, the model quickly learns to allocate the minimum budget, keeping response length stable during RL training.

An alternative strategy infers problem difficulty from performance metrics rather than direct prediction. ALP~\cite{xiang2025justthinkingefficientreasoning}, for example, serves the per-prompt solve rate as a proxy for difficulty, applying a higher length penalty to easier tasks (i.e. those with high solve rates). Similarly, ACPO~\cite{cheng2025incentivizingdualprocessthinking} uses the sampling success rate to incentivize effective allocation of fast and slow thinking modes. When the rate exceeds a specified threshold, the reward encourages fast thinking; when it is below the threshold, it promotes slow thinking.

In a notable departure, some work reverses this logic, using response length as an implicit signal of difficulty. Rather than using a difficulty estimate to control length, \citet{ling2025fasteasydeephard} hypothesizes that the length of a successfully generated reasoning chain is itself a reliable indicator of the inherent complexity of the problem. This allows them to use response length as an implicit proxy for difficulty when shaping rewards.

\subsubsection{RL with Thinking Mode}
\label{sec:rl_with_think_modes}

Several lines of work use RL to explicitly explore models to switch between different ``thinking modes'', most commonly a deliberative ``Thinking'' mode (e.g., CoT) and a reactive ``No Thinking'' mode (direct answer). The primary distinction lies in how mode selection is framed and optimized.

The most direct approach is to train a policy to explicitly select a reasoning mode. For instance, AdaptThink~\cite{zhang2025adaptthinkreasoningmodelslearn} employs RL to enable reasoning models to adaptively choose between ``Thinking'' and ``NoThinking'' modes based on problem difficulty. The algorithm combines a penalty-augmented objective to maximize ``NoThinking'' usage under accuracy constraints with an importance sampling strategy that ensures balanced exploration during training.  Similarly, \citet{jiang2025thinkneedlargehybridreasoning} aim to develop a hybrid thinking system that balances system 2 reasoning with system 1 capabilities. They sample using both ``think'' and ``no think'' models to create diverse output candidates, applying a rule-based scheme for inter-group and intra-group rewards. This standardized reward mechanism helps the model effectively learn adaptive reasoning mode selection strategies.

Other methods achieve adaptive behavior through more nuanced or emergent mechanisms. Autothink~\cite{tu2025learningthinkshapingadaptive}, for example, triggers stochastic mode selection by inserting special tokens (``...'') and then uses a multi-stage RL process to stabilize this behavior and optimize for conciseness. On the other hand, \citet{lou2025adacotparetooptimaladaptivechainofthought} model adaptive CoT triggering as a Pareto optimization problem, using PPO algorithm to dynamically adjust penalty coefficients and control the decision boundary between CoT and direct-answer modes, balancing performance and computational cost.

\subsubsection{Other RL}
\label{sec:rl_other}

Some strategies are distinguished by their reward designs, training frameworks, or conceptual approaches to reasoning efficiency.

There are some works which contribute to the design of the reward function, moving beyond simple length penalties. Some learn the reward dynamically; ConciseRL~\cite{dumitru2025conciserlconcisenessguidedreinforcementlearning} converts semantic conciseness into a learnable reward by using an LLM as a dynamic judge to evaluate reasoning compactness in terms of logical necessity and redundancy. It introduces two reward functions: a pure conciseness reward and an accuracy-gated variant to promote concise and accurate reasoning while minimizing evaluation cost. Others use relative comparisons; TWYN~\cite{yang2025thinkneedselfadaptivechainofthought} adopts a pairwise comparison-based reward mechanism, which computes rewards based on the relative relationships between samples rather than absolute metrics, aiming to jointly optimize reasoning accuracy and response conciseness in LLMs' generation. L2T~\cite{wang2025learningthinkinformationtheoreticreinforcement} takes LLM reasoning optimization as an episodic RL task, employing a dense process reward to promote efficient inference under token constraints. This reward integrates information gain from model updates with a parameter compression penalty derived from PAC-Bayes theory and Fisher information, which is efficiently approximated through low-rank methods. A different approach is seen in \citet{xie2025interleavedreasoninglargelanguage}, which employs interleaved reasoning via RL, enabling LLMs to alternate between thinking and answering in multi-hop tasks. It introduces conditional intermediate rewards—applied only during stable training phases—under all-or-nothing, partial credit, and time-discounted schemes, with the latter encouraging earlier correct steps.

Another set of innovations comes from novel training frameworks and schedules. For example, \citet{fatemi2025concisereasoningreinforcementlearning} propose a two-stage RL framework to improve the efficiency of LLM reasoning. They train the model on hard problems in the first stage, allowing response length expansion to explore complex solutions under negative reward signals. Continually, they optimize conciseness on small solvable sets via PPO in the second stage, rewarding shorter correct output while preserving exploration to avoid overfitting. Long\(\otimes\)Short~\cite{ning2025thoughtsgeneratedequalefficient} synergistically combines a long-thought LLM and a short-thought LLM to collaboratively enhance reasoning capabilities: the former specializes in generating essential insights, while the latter efficiently produces supplementary ideas. The approach also employs asynchronous strategy optimization to facilitate dynamic switching between models, further improving collaboration through a hybrid reward system that emphasizes correctness, format, and length.

Finally, some works introduce new high-level concepts to guide optimization. \citet{yi2025shorterbetterguidingreasoningmodels} propose the concept of an inherent ``Sample Optimal Length'' (SOL)—the shortest correct output for a given problem—and use it as a dynamic reward signal to allow the model to autonomously learn its own optimal reasoning length. In a similar vein, \citet{gao2025faroptimalreasoningefficiency} introduce the ``Reasoning Efficiency Gap'' (REG) as a metric to quantify the combined cost of token waste and accuracy loss. They then propose the REO-RL algorithm to directly minimize this gap by strategically optimizing for a sparse set of token budgets.

\subsection{Summary}
\label{sec:training_based_summary_anf_outlook}

Training-based methods provide critical support to enhance the adaptive reasoning capabilities of LLMs. We survey training-based approaches mainly including fine-tuning and RL based methods. We first review the data construction methods for variable-length data and collect the existing open-source variable-length datasets for reasoning. In terms of fine-tuning methods, we meticulously categorize the methods into five classes and perform a detailed analysis and summary of each. Long CoT compression methods improve reasoning efficiency and adaptability but face trade-offs between compression effectiveness and reasoning fidelity, along with increased data requirement and generalization challenges; while short CoT selection fine-tuning improves reasoning efficiency by promoting concise or self-verified reasoning paths, but may risk omitting critical steps or requiring complex training procedures and careful balancing between brevity and accuracy; implicit CoT fine-tuning achieves efficiency through latent reasoning representations or knowledge distillation but sacrifices interpretability due to non-explicit reasoning steps and risks misalignment between compressed representations and task requirements; DPO variant approaches enable multi-objective optimization balancing conciseness and accuracy via preference learning, yet face challenges in constructing high-quality preference pairs and maintaining reasoning depth under strict length constraints; other hybrid methods combine fast/slow cognitive systems or novel loss functions to achieve adaptive reasoning, though they often require complex training pipelines and careful calibration of dual-mode interactions.

RL methods balance conciseness and accuracy through five key paradigms. RL with Length Penalty improves efficiency by penalizing verbose outputs via reward shaping or external constraints, but risks oversimplification of complex tasks or overfitting to penalty thresholds. GRPO-Variant approaches address ``Format Collapse'' by diversifying reasoning modes or integrating difficulty-aware rewards, though they often require intricate loss designs and multi-component systems. Difficulty-Aware RL adapts response length to problem complexity through explicit difficulty estimation or implicit signals (response length, solve rates), yet faces challenges in accurate difficulty calibration and generalization across domains. Thinking Mode RL enables dynamic switching between deliberative (``Thinking'') and reactive (``No Thinking'') modes, but struggles with mode selection stability and exploration-exploitation trade-offs. Other RL Innovations introduce learnable reward functions, hybrid frameworks, or novel metrics, though these often demand extensive computational resources or face scalability issues.

\section{Challenges and Future Directions}
\subsection{Model Capability-aware Reasoning}
Early research on concise thinking aims to enforce strict length constraints on reasoning outputs through unified principles like ``Be Concise'' or the hard token budget. Given the prevalent overthinking tendencies in current open-source LRMs and their narrow evaluation benchmarks, this approach does achieve observable performance by reducing reasoning length without significant performance degradation on mathematics and programming tasks. However, this methodology exhibits clear limitations. While simple queries might permit near-instant responses without thinking, complex problems like Olympiad-level mathematics or programming tasks inherently demand extended reasoning chains. Consequently, subsequent studies increasingly incorporate input difficulty awareness to enable adaptive thinking. Yet relying solely on human-prioritized input difficulty is insufficient. Given a problem, different LRMs may produce varying output lengths due to disparities in their base models' capacity. For example, some SFT-based approaches use variable-length CoT data, either from open-source datasets or curated with well-designed pipelines. Such CoT data may well reflect the input difficulty, but it may not accurately reflect the model's inherent knowledge and problem-solving abilities for questions of varying complexity, that is, how many tokens the model needs for thinking to produce an answer. This underscores the critical need to optimize adaptive thinking by simultaneously integrating input difficulty with model capability calibration.


\subsection{Human Preference-aware Reasoning}
The significant impact of LRMs stems not only from their exceptional performance on complex reasoning tasks but also from their ability to demonstrate explicit reasoning processes, which substantially enhances answer interpretability. These thinking processes often provide users with novel insights~\cite{hammoud2025answerreasoningtraceuncovers}. From this perspective, many existing concise thinking or adaptive thinking methods contradict this fundamental strength. Aggressively simplifying steps, penalizing output length, or compressing explicit reasoning into latent space risks reducing the model's interpretability. We therefore argue that in advancing concise or adaptive thinking research, beyond input difficulty and model capacity, human preference must be considered. For common-sense domains or questions widely recognized as self-explanatory within real user query distribution, concise thinking may be appropriate. However, in specialized fields like medical, financial, or legal domain, even for seemingly straightforward queries, simplifying the thinking steps could undermine interpretability, particularly when users may lack relevant background knowledge. Future research needs to establish standards for good reasoning chains across various domains, ensuring that models' reasoning processes align closely with human preference. CoTs that conform to human cognitive patterns, even when extended in length, can enhance answer comprehension.

\subsection{Trustworthy Reasoning} The reliability of LRMs, particularly regarding hallucination and safety~\cite{deepseekr1hallucinate,o3truthfulness}, remains one of the most critical challenges for real-world deployment.

For hallucination, while some studies~\cite{cheng2025thinkmorehallucinateless} suggest that enabling models to engage in step-by-step reasoning can enhance their ability to reflect and utilize internal knowledge, thereby reducing hallucinations, this approach may introduce a new type of hallucination known as reasoning hallucination~\cite{sun2025detectionmitigationhallucinationlarge,yao2025reasoningmodelspronehallucination} or overconfident with deeper reasoning~\cite{mei2025reasoninguncertaintyreasoningmodels}. Such hallucinations emerge when reasoning processes maintain structural integrity but contain factual inaccuracies, producing conclusions that appear convincing at surface level yet ultimately incorrect.

For safety, there is no widely accepted conclusion regarding the impact of response length on safety, or whether long CoT actually enhances the safety performance of LLMs. Some works~\cite{li2025outputlengtheffectdeepseekr1s,liu2025selfreflectionmakeslargelanguage,wang2025star1saferalignmentreasoning} show that longer CoT reasoning can improve safety via deliberative reasoning and self-correction. So shortening the reasoning length could probably lead to performance degradation on safety. While ~\citet{zhao2025tradeoffslargereasoningmodels,jiang2025safechainsafetylanguagemodels} also reveal that long thoughts are more unsafe than answers, there exists a trade-off between reasoning and safety capability~\cite{huang2025safetytaxsafetyalignment}. \citet{zhang2025enhancesafetylargereasoning} find that simply using short or template-based reasoning processes can attain comparable safety performance. \citet{zhang2025continuethinkingadaptivethinking} show that after adaptive thinking is applied to LRMs, the safety performance is significantly improved since the long CoT is only triggered when necessary. \citet{zhu2025thinkthinkexploringunthinking} expose that the thinking process of LRMs can be entirely bypassed through simple token manipulations (e.g., modifying chat template). Such ``unthinking vulnerability'' can be used in backdoor attack of LRMs and hurt their reliability.

Beyond hallucination and safety, another fundamental capability that could impact LRMs' reliability is instruction following. While LRMs have demonstrated human-level intelligence across numerous challenging benchmarks, a concerning trend has emerged: the smarter the models become, the less compliant they are; in other words, their ability to follow instructions diminishes. \citet{fu2025scalingreasoninglosingcontrol} identify a trade-off between reasoning and instruction-following ability. Improving reasoning capability often comes at the cost of instruction adherence. \citet{zhu2025largereasoningmodelssave,zhu2025thinkthinkexploringunthinking} point out that LRMs can be attacked to skip thinking or re-engage thinking process when the think process has already been completed. There are still issues of reasoning reliability of LRMs.

The aforementioned three capabilities are crucial for the reliability and real-world application of LRMs. However, current research on LRMs put little attention on these aspects, particularly in scenarios involving concise thinking or adaptive thinking, where few studies have systematically considered or evaluated these three capabilities. When defining good reasoning chains in future research, it is essential to incorporate these fundamental model capabilities.

\bibliography{references}

\end{document}